%% file: main.tex
\documentclass[acmtog]{acmart}
\acmSubmissionID{760}

\usepackage[ruled]{algorithm2e} 

\SetAlFnt{\small}
\SetAlCapFnt{\small}
\SetAlCapNameFnt{\small}
\SetAlCapHSkip{0pt}

\acmJournal{TOG}

\usepackage{url}            
\usepackage{amsfonts}       
\usepackage{nicefrac}       
\usepackage{stfloats}
\usepackage{multirow}
\usepackage{makecell}
\usepackage{enumitem}

\usepackage[capitalize]{cleveref}
\crefname{section}{Sec.}{Secs.}
\Crefname{section}{Section}{Sections}
\Crefname{table}{Table}{Tables}
\crefname{table}{Tab.}{Tabs.}

\newcommand{\ie}{\textit{i}.\textit{e}.}
\newcommand{\eg}{\textit{e}.\textit{g}.}

\newcommand{\wrt}{\textit{w}.\textit{r}.\textit{t}.}
\newcommand{\nickname}{MV-S2V}

\copyrightyear{2026}
\acmYear{2026}
\setcopyright{cc}
\setcctype{by-nc-nd}
\acmConference[SIGGRAPH Conference Papers '26]{Special Interest Group on Computer Graphics and Interactive Techniques Conference Conference Papers}{July 19--23, 2026}{Los Angeles, CA, USA}
\acmBooktitle{Special Interest Group on Computer Graphics and Interactive Techniques Conference Conference Papers (SIGGRAPH Conference Papers '26), July 19--23, 2026, Los Angeles, CA, USA}
\acmDOI{10.1145/3799902.3811131}
\acmISBN{979-8-4007-2554-8/2026/07}

\begin{document}
\title{\nickname: Multi-View Subject-Consistent Video Generation}

\author{Ziyang Song}
\affiliation{
 \institution{The Hong Kong Polytechnic University}
 \country{China}}
 \email{ziyang.song@connect.polyu.hk}
\author{Xinyu Gong$^{\dagger}$}
\affiliation{
 \institution{The University of Texas at Austin}
 \country{USA}}
 \email{xinyu.gong@utexas.edu}
\author{Bangya Liu}
\affiliation{
 \institution{University of Wisconsin-Madison}
 \country{USA}}
 \email{bliu277@wisc.edu}
\author{Zelin Zhao}
\affiliation{
 \institution{Georgia Institute of Technology}
 \country{USA}}
 \email{zelin@gatech.edu}

\thanks{$^{\dagger}$Corresponding author: Xinyu Gong.}

\input{sec/0_abstract}

%
%


%
%


\begin{teaserfigure}
  \includegraphics[width=1.0\textwidth]{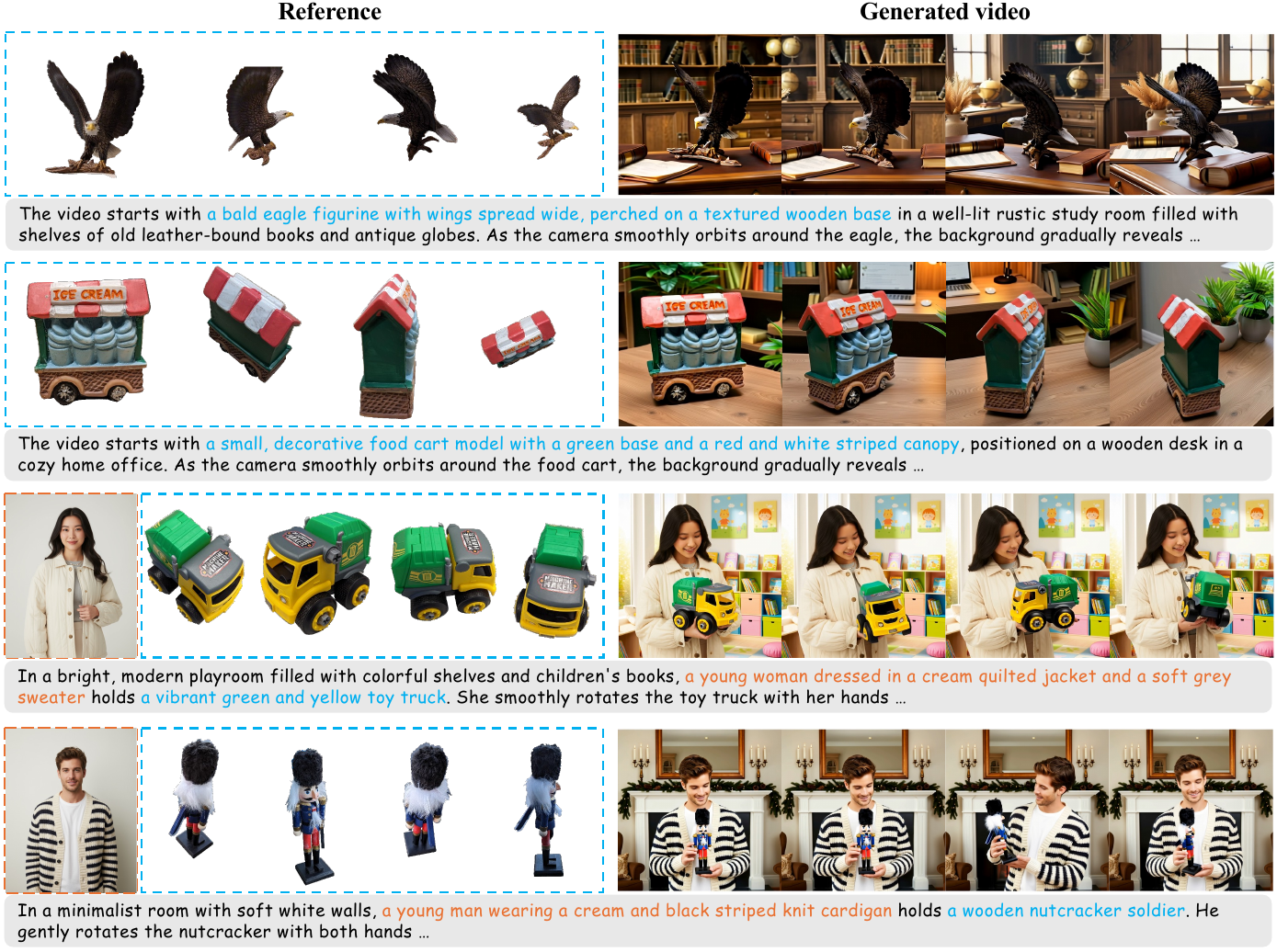}
  \vspace{-0.5cm}
  \caption{Given multi-view reference images for subjects, our \nickname{} can generate videos with multi-view (3D) subject consistency.}
  \label{fig:overview}
\end{teaserfigure}

\maketitle

\input{sec/1_intro}
\input{sec/2_liter}
\input{sec/3_data}
\input{sec/4_method}
\input{sec/5_expm}
\input{sec/6_sum}

\clearpage
\begin{figure*}[t]
\centering
  \includegraphics[width=0.93\linewidth]{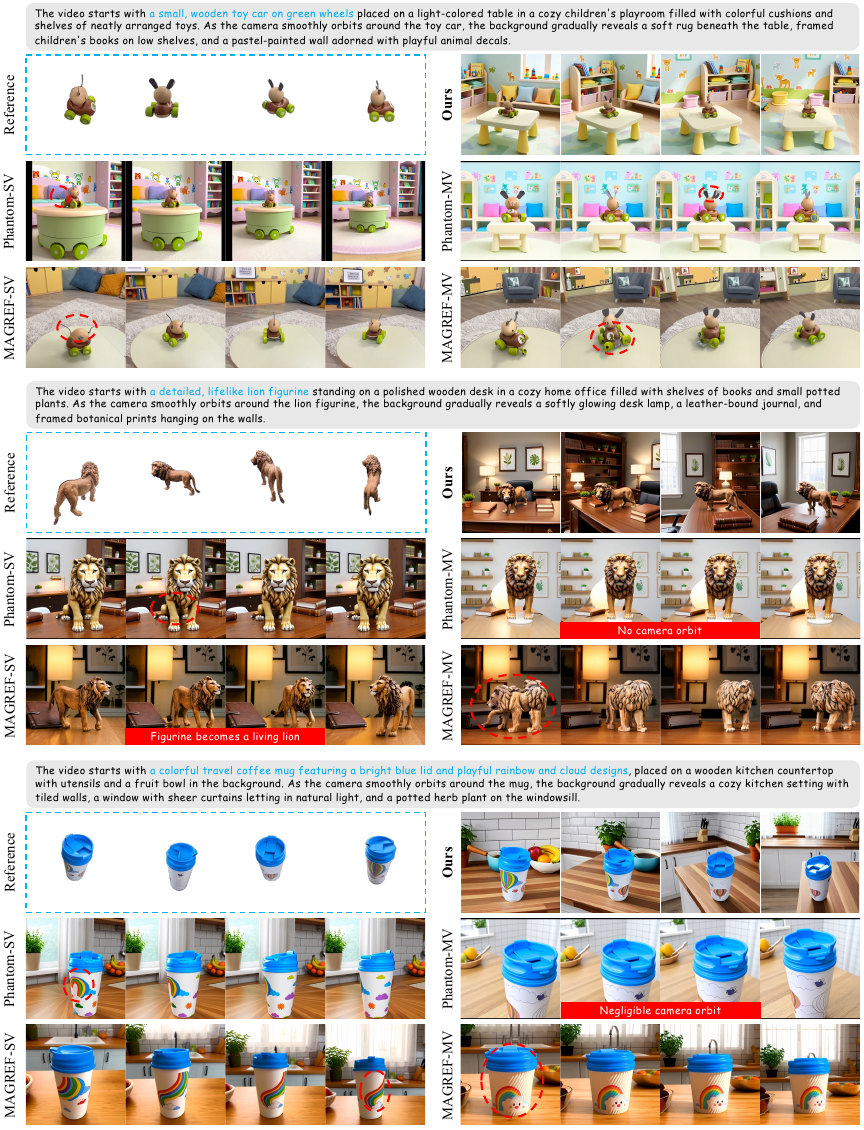}
  \vspace{-0.5cm}
  \caption{Qualitative results of all methods on Object-Centric (OC) scenes. Inconsistencies and artifacts in generated results are highlighted.}
  \label{fig:qual_res1}
\end{figure*}

\begin{figure*}[t]
\centering
  \includegraphics[width=0.92\linewidth]{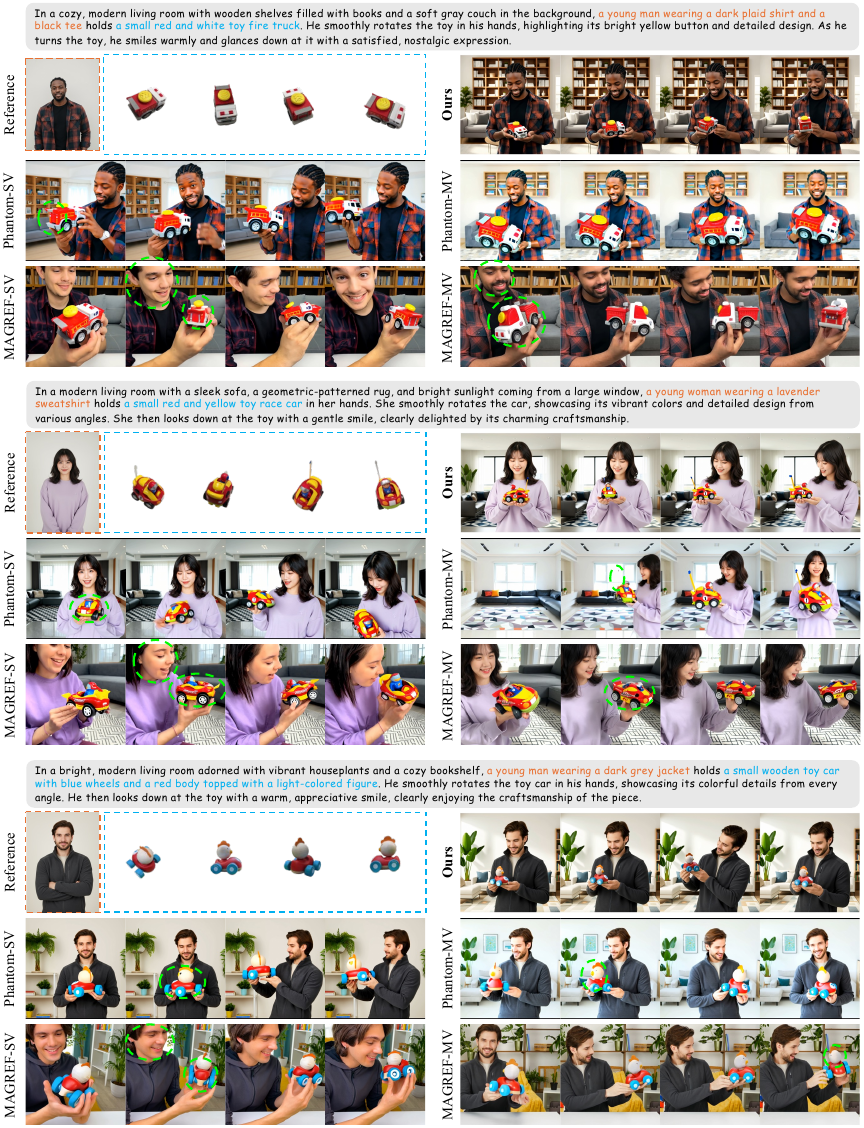}
  \vspace{-0.5cm}
  \caption{Qualitative results of all methods on Human-Object Interaction (HOI) scenes. Inconsistencies and artifacts in generated results are highlighted.}
  \label{fig:qual_res2}
\end{figure*}

\clearpage
\bibliographystyle{ACM-Reference-Format}
\bibliography{main}

\clearpage
\appendix
\input{sec/X_suppl}

\end{document}

%% file: sec/0_abstract.tex
\begin{abstract}
Existing Subject-to-Video Generation (S2V) methods have achieved high-fidelity and subject-consistent video generation, yet remain constrained to single-view subject references. This limitation renders the S2V task reducible to an S2I + I2V pipeline, failing to exploit the full potential of video subject control. In this work, we propose and address the challenging Multi-View S2V (MV-S2V) task, which synthesizes videos from multiple reference views to enforce 3D-level subject consistency. Regarding the scarcity of training data, we first develop a synthetic data curation pipeline to generate highly customized synthetic data, complemented by a small-scale real-world captured dataset to boost the training of MV-S2V. Another key issue lies in the potential confusion between cross-subject and cross-view references in conditional generation. To overcome this, we further introduce Temporally Shifted RoPE (TS-RoPE) to distinguish between different subjects and distinct views of the same subject in reference conditioning. Our framework achieves superior 3D subject consistency \wrt{} multi-view reference images and high-quality visual outputs, establishing a new meaningful direction for subject-driven video generation. Code and data are available at \url{https://szy-young.github.io/mv-s2v}
\end{abstract}

%% file: sec/1_intro.tex
\section{Introduction}
\label{sec:intro}

The video generation landscape has been fundamentally reshaped by the technical maturity of diffusion models \cite{Ho2020,Peebles2023}. This progress has successfully enabled the creation of high-quality videos from diverse inputs, most notably through Text-to-Video (T2V) \cite{Sora} and Image-to-Video (I2V) \cite{Blattmann2023} frameworks. Building on this, Subject-to-Video generation (S2V) \cite{Huang2025,Chen2025b} has emerged. S2V takes text prompt and a set of reference images for main subjects as inputs and enforces identity consistency for the subjects across the generated video, offering greater controllability than T2V and higher flexibility than I2V. 

However, the S2V paradigm faces two critical limitations. First, high-quality data collection is notoriously costly \cite{Liu2025,Chen2025c,Zhang2025}. Second, current S2V methods typically take in only a single reference image for each subject, thereby controlling the subject appearance of only a single view in the generated video through reference conditioning. The S2V framework under such single-view setting can be readily decomposed into a pipeline of Subject-to-Image (S2I) followed by Image-to-Video (I2V), while training data for the two sub-tasks are much simpler to acquire than for S2V. This naturally leads to the question: what are the fundamental advantages of S2V?

In this work, we commit to a more ambitious while practical goal: \textbf{Multi-View Subject-to-Video Generation (MV-S2V)}. Specifically, given multiple reference images capturing a subject from different views, our goal is to synthesize a video where the subject adheres to multi-view subject consistency with reference images. We argue that this multi-view subject control represents the core value of S2V that truly differentiates it from the "S2I + I2V" pipeline, \ie, to utilize reference images from various views or states to comprehensively control the dynamic appearance of subjects throughout the video. Besides, the formulated multi-view S2V task holds significant values for real-world applications requiring high fidelity to subjects, such as advertising and augmented reality.

This ambitious goal of multi-view S2V faces two main challenges. The first challenge is the lack of suitable training data. Multi-view S2V expects training videos which showcase the subjects from diverse views. However, such videos are not prevalent in massive web video data, making direct curation infeasible. To address this, we construct a highly controllable synthetic data curation pipeline: We leverage the camera controllability and prompt following ability of existing I2V methods \cite{Cao2025,Wang2025b} to customize the generation of a large volume of videos featuring multi-view subject showcases, from which we can also extract the corresponding multi-view reference images. Simultaneously, to mitigate the "copy-paste" effects possibly introduced by directly extracting reference images from videos, we capture a small-scale S2V dataset in the real world, where the videos and multi-view reference images are entirely decoupled. The joint utilization of these two data sources enables the model to grasp the multi-view conditioning capability from diverse, large-scale data while improving its robustness to arbitrarily captured real-world images.

The second challenge lies in reference conditioning. When extending from single-view to multi-view S2V, it is crucial to further distinguish between different subjects and distinct views of the same subject in reference conditioning. The conditioning mechanism in existing methods, \eg, concatenating references along the frame dimension \cite{Jiang2025,Liu2025}, or compositing references on a single image \cite{Deng2025}, fail to distinguish these two cases. To address this, we propose a tailored reference conditioning mechanism, \textbf{Temporally Shifted RoPE (TS-RoPE)}, which clearly separates different subjects and views via rotary position encoding (RoPE).

In summary, our contributions are four-fold:

\begin{itemize}[leftmargin=*]
\setlength{\itemsep}{1pt}
\setlength{\parsep}{1pt}
\setlength{\parskip}{1pt}
\item \textbf{Formulation:} We formulate the Multi-View Subject-to-Video Generation (MV-S2V) task, highlighting the core value of S2V paradigm over a sequential S2I+I2V pipeline.
\item \textbf{Data:} We introduce a data curation pipeline to boost MV-S2V training with customized high-quality training data.
\item \textbf{Method:} We propose TS-RoPE which effectively distinguishes between cross-subject and cross-view references in conditioning.
\item \textbf{Evaluation:} We design a series of evaluation metrics to measure multi-view and 3D subject consistency. Extensive experiments demonstrate the superior performance of our approach on such high-fidelity subject consistency.
\end{itemize}

%% file: sec/2_liter.tex
\section{Related Work}
\label{sec:liter}

\begin{figure*}[t]
\centering
  \includegraphics[width=1.0\linewidth]{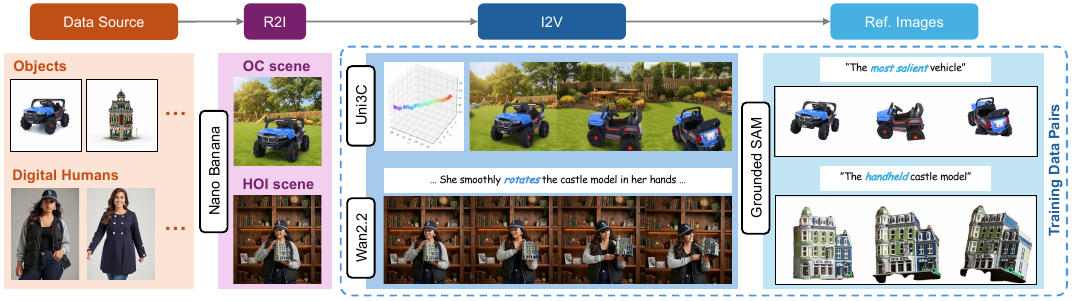}
  \vspace{-0.6cm}
  \caption{Synthetic data curation pipeline for \nickname, where the use of existing I2V models enables highly customized training data generation. Video captioning and data filtering stages are omitted for brevity.}
  \label{fig:data}
\end{figure*}

\subsection{Video Foundation Models}

The advancement of diffusion models has significantly accelerated the research and development of video foundation models, yielding impressive content creation and intelligent interaction. Early methods, \eg, Stable Diffusion 1.5 \cite{Rombach2O22}, are mainly based on latent diffusion models (LDMs) \cite{Rombach2O22} with a U-Net architecture \cite{Ronneberger2015}. Such models were later augemented with temporal modules for video generation, leading to models such as Make-A-Video \cite{Singer2023}, SVD \cite{Blattmann2023}, and AnimateDiff \cite{Guo2024}. A pivotal architectural shift came with Diffusion Transformer (DiT) \cite{Peebles2023}, which applied scaling laws to generative models and resulted in powerful models like Wan \cite{Wang2025b}. The MMDiT, featuring a dual-stream DiT architecture, was meanwhile introduced in Stable Diffusion 3 \cite{Esser2024} and later adopted by leading open-source video generation projects including CogvideoX \cite{Yang2025}, HunyuanVideo \cite{Kong2024}, and SeedVR \cite{Wang2025}.

\subsection{Subject-Consistent Image Generation}

Early progress in subject-consistent image generation relies on optimization-based methods \cite{Hu2022,Huang2024,Shah2024,Gal2023,Ruiz2023} that train identifiers to bind image content. A significant training-based approach is IP-Adapter \cite{Ye2023}, which achieves consistency by freezing the base model and training specialized adapters only. While adapters are popular for tasks like facial ID consistency \cite{Wang2024,Guo2024b,Chen2024}, their reliance on CLIP \cite{Cherti2023} or DINO \cite{Oquab2024} features often creates a trade-off between detail preservation and prompt following ability. PuLID \cite{Guo2024a} introduces contrastive alignment to resolve this trade-off, enabling efficient yet precise ID customization. Omni-ID \cite{Qian2025} proposes a holistic identity representation to capture full subject attributes, addressing the limitation of narrow ID focus in existing adapter-based schemes. A newer trend integrates generation and editing into a unified framework \cite{Chen2025,Han2025,Xiao2025}. Unlike adapter methods, this approach better leverages foundation models to learn image-text alignment, avoiding the performance degradation often caused by using multiple adapters. 

\subsection{Subject-Consistent Video Generation}

Optimization-based methods like Kling \cite{Kling} address video identity consistency by requiring multiple user-uploaded videos for fine-tuning, which is computationally expensive. Meanwhile, adapter-based approaches such as ID-Animator \cite{He2024} and ConsisID \cite{Yuan2025} have emerged as alternatives. However, these methods are often evaluated on small datasets ($\sim$10k samples), limiting their generalization and ability to align detailed subject features with text descriptions. While recent works \cite{Huang2025,Liang2025,Chen2025,Jiang2025,Liu2025,Deng2025,Zhang2025,Zhao2025} have demonstrated consistent video generation with multiple subjects, they remain limited to single-view references and fail to fully exploit subject control capabilities of video generation. The concurrent work \cite{Liu2025c} also explores 3D consistency \wrt reference subjects in video generation, while it requires per-frame 6DoF subject poses in the generated video as input, comprimising its flexibility and usability. 

\subsection{3D Generation and Novel View Synthesis}

Recent advances in diffusion-based 3D generation and novel view synthesis (NVS) have enabled strong 3D-aware content creation from multi-view cues. Methods such as Zero123 \cite{Liu2023Zero123}, SyncDreamer \cite{Zhou2024SyncDreamer}, MVDream \cite{Shi2023MVDream}, and SV3D \cite{Voleti2024SV3D} learn multi-view consistent priors and can naturally accommodate multi-view references as conditional inputs, producing geometrically coherent novel views for static objects and scenes. Large-scale 3D generative models including LGM \cite{Tang2024LGM} and SLAT \cite{Xiang2025SLAT} further improve fidelity and scalability by leveraging large 3D asset datasets, yet they are still heavily constrained by the limited scale of 3D training data and cannot directly generate dynamic scenes.

\subsection{RoPE Manipulation in Diffusion Models}

Since transformers lack spatial awareness, modern DiT models adopt rotary positional encodings (RoPE) \cite{Su2024} to encode relative positions. Recently, some works have employed RoPE to inject various inductive biases into the DiT architecture. Qwen-Image \cite{Wu2025} proposes Multimodal Scalable RoPE (MSRoPE) for better image resolution scaling and improved text-image alignment. AlignedGen \cite{Zhang2025a} introduces ShiftPE to address positional collisions in an attention-sharing framework for style-aligned image generation. PE-Field \cite{Bai2026} models spatial correspondence via RoPE for novel view synthesis. MinT \cite{Wu2025a} designs a temporal-aware ReRoPE to guide video generation with temporal event control. In this paper, we also manipulate RoPE to address the confusion between cross-subject and cross-view references.

%% file: sec/3_data.tex
\section{Dataset Construction}
\label{sec:data}

To successfully train a multi-view S2V model, a dedicated dataset is essential, requiring \textit{(video, references, text)} data triplets. Especially, we expect that the video explicitly displays the different sides of subjects, establishing a correspondence with the multi-view reference images. In this work, we focus on two typical types of videos featuring multi-view subject showcases: \textbf{1) Object-Centric (OC):} Camera orbiting videos that display the static central objects from different perspectives through camera movements. \textbf{2) Human-Object Interaction (HOI):} Videos where persons manipulate the hand-held objects to display their different sides.

However, videos that naturally showcase subjects from multi-views are scarce among the vast volume of web videos. Directly mining web video data brings substantial computational and memory cost, yet yields a low proportion of usable samples. Some existing OC datasets, e.g., Co3D \cite{Reizenstein2021}, provide videos with orbiting camera trajectories, while the camera movements in the videos are highly jittering and both the subjects and backgrounds lack diversity, thus degrading the smoothness, diversity, and visual quality of generated videos. While existing HOI datasets, e.g., HOIGen-1M \cite{Liu2025b}, reach an impressive data volume, the proportion of videos demonstrating multiple object views remains negligibly small.

\subsection{Synthetic Data Curation} 

To overcome the data scarcity, we seek to synthetic data source via existing Image-to-Video (I2V) generation models, motivated by two key facts: 1) I2V models have advanced rapidly, leading to generated videos with high visual quality. 2) Unlike real-world videos, I2V-generated content can be controlled through various conditioning, enabling highly customized training data generation. In this regard, we introduce the following multi-stage synthetic data curation pipeline. Figure \ref{fig:data} illustrates our whole data pipeline.

\textbf{(1) Image Synthesis.} We start by composing our internal collection of object and human asset images into full scene images via the Subject-to-Image (S2I) model Nano-Banana \cite{Banana}. The initial data source contains $\sim$ 16,000 objects, primarily covering four categories: \textit{Beauty \& Personal Care}, \textit{Shoes}, \textit{Luggage \& Bags}, and \textit{Toys \& Hobbies}, with a 1:1:1:7 ratio. Here \textit{Toys \& Hobbies} occupies a major part given its greater diversity in shape and appearance than the other three categories. We also include 4,734 human images with balanced distribution in gender (female/male), age (young/middle-aged), and race (Asian/Caucasian/African/Hispanic). Each object asset leads to an OC image and an HOI image, with a human subject randomly sampled for composing the HOI scene.

\textbf{(2) Video Synthesis.} With the high-fidelity images, we proceed to video generation using I2V models. For Object-Centric (OC) videos, we employ Uni3C~\cite{Cao2025}, a camera-controllable video generation model, which allows us to explicitly control the camera's trajectory and thereby guarantee a multi-view object display throughout the video. For Human-Object Interaction (HOI) videos, we select Wan2.2 \cite{Wang2025b} due to its exceptionally strong prompt following ability, ensuring the person in the generated video accurately executes the desired sequence of actions, \ie, smoothly rotating the object to reveal its multiple facets.

\textbf{(3) Video Captioning.} We employ Taiser2 \cite{Yuan2025b} to generate high-quality textual descriptions for each training video. We also generate word descriptions about the main subjects, \ie, the central object in an OC scene or the handheld object in an HOI scene, for the following reference extraction step.

\textbf{(4) Reference Extraction.} From the generated videos above, we sample key frames and employ Grounded SAM \cite{Ren2024} to segment and crop out the main subjects, forming multi-view references. However, it is non-trivial to uniquely identifying the desired subjects from these synthetic videos. Due to scene complexity and presence of multiple object instances in the videos, simple category-level word descriptions (\eg, \textit{book}, \textit{figurine}) often cause Grounded SAM to output multiple detections, while overly detailed prompts tend to confuse the model. We find a simple yet robust strategy by simply prepending focus-driven modifiers to the prompt, \eg, \textit{the \textbf{most salient} book}, \textit{the \textbf{handheld} figurine}. This straightforward prompt augmentation improves the segmentation usability from 15\% to over 90\% without any post-processing.

To reduce the "copy-paste" effect caused by extracting references from videos, we generate relatively long raw videos in the previous I2V step and clip shorter segments as training videos, while reference key frames are still taken from raw videos. In this way, a part of reference views are fully decoupled from the training videos. Data augmentations on scale, rotation, shift, and brightness are also applied to the references. While some methods apply S2I to augment the poses for objects, we avoid doing so as it may introduce inconsistency among multiple reference views.

\textbf{(5) Data Filtering.} An advanced Vision-Language Model, \ie, Gemini 2.5 \cite{Gemini}, is employed to automatically prune low-quality data in synthesized videos, \eg, human body artifacts, physically implausible floating objects, and distracting elements like subtitles or watermarks.

In total, we collect 11,804 and 10,130 training samples for OC and HOI scenarios, respectively.

\begin{figure}[t]
\centering
  \includegraphics[width=1.0\columnwidth]{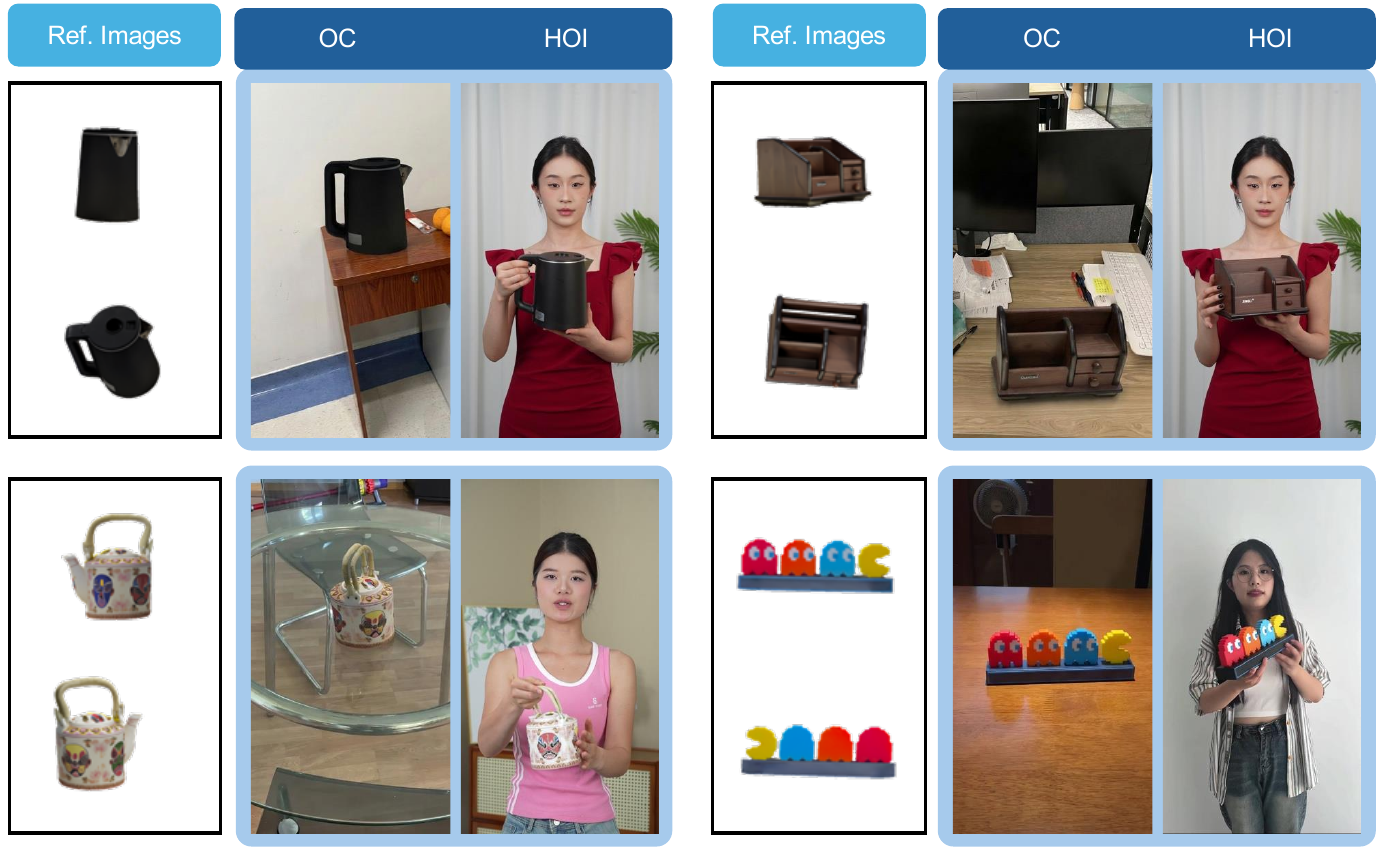}
  \vspace{-0.6cm}
  \caption{Examples of real-world dataset.}
  \vspace{-0.6cm}
  \label{fig:data_real}
\end{figure}

\subsection{Real-world Data Capture} 

To further enhance the photorealism and generalizability of our models, we complement our synthetic dataset with a small-scale real-world dataset. In this dataset, videos and multi-view reference images are captured separately for both OC and HOI scenarios. This capture process fully decouples the object poses in training video from those in reference images, further mitigate the "copy-paste" effect. We use 100 distinct objects, with 5 young Asian females acting in HOI data capture. In total, we collect 1,724 and 1,514 training samples for OC and HOI scenarios, respectively. Examples of this dataset are shown in Figure \ref{fig:data_real}.

%% file: sec/4_method.tex
\section{\nickname}
\label{sec:method}

\begin{figure*}[t]
\centering
  \includegraphics[width=1.0\linewidth]{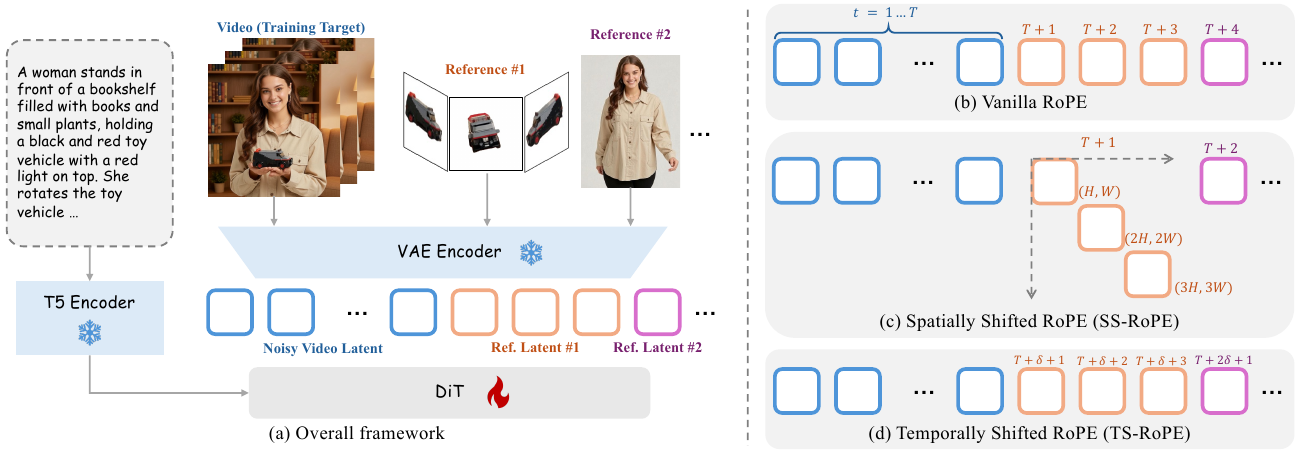}
  \vspace{-0.6cm}
  \caption{Illustration about our MV-S2V framework along with different designs for multi-view reference conditioning.}
  \label{fig:method}
\end{figure*}

\subsection{Preliminary: T2V Base Model}

We aim to build a framework which integrates multi-view reference images of subjects into video diffusion process. Specifically, the input conditions include a textual description $y$ and a set of reference images $\boldsymbol{R} = \{\boldsymbol{R}_1, \boldsymbol{R}_2, ...\}$, where $\boldsymbol{R}_i = \{I^{r_i}_1, ..., I^{r_i}_{M_i}\}$ denotes $M_i$ reference views of the $i$-th subject. Our goal is to generate a $T_0$-frame video $\boldsymbol{V} = \{I^v_1, ..., I^v_{T_0}\}$ from the inputs, and the overall objective equals to the modeling of the following conditional distribution:
\begin{equation}
    p(\boldsymbol{V} | \boldsymbol{R}, y) = p(I^v_1, ..., I^v_{T_0} | I^{r_1}_1, ..., I^{r_1}_{M_1}; I^{r_2}_1, ..., I^{r_2}_{M_2}; ...; y)
\end{equation}

Our framework is built upon a pre-trained text-to-video foundation model, Wan 2.1~\cite{Wang2025b}. As shown in Figure \ref{fig:method} (a), the input head consists of a 3D Variational Auto-Encoder (VAE), which compresses the $T_0$-frame target video $\boldsymbol{V}$ into a latent feature tensor $F^v \in \mathbb{R}^{T \times C \times H \times W}$, where $T$ and $H \times W$ denote the temporal length and spatial resolution after compression respectively, and $C$ refers to the feature channel dimension. The reference images also share the 3D VAE encoder for feature extraction, leading to latent space alignment for visual inputs. Specifically, each reference image $I^{r_i}_{m_i}$ is independently processed into a reference feature tensor $F^{r_i}_{m_i} \in \mathbb{R}^{C \times H \times W}$. A DiT network \cite{Peebles2023} iteratively denoises the data on this latent space. The textual input $y$ is encoded through a T5 encoder \cite{Raffel2019} and fused with visual features through cross-attention layers.

\subsection{Multi-View Reference Conditioning}
\label{sec:rope}

Reference conditioning in video generation typically adopts either adapter modules \cite{He2024,Yuan2025} or self-attention mechanisms \cite{Jiang2025,Liu2025}, with the latter proved to be more effective at subject consistency and detail preservation. Our method also adopts the simple yet effective self-attention-based reference conditioning. Specifically, video latents $F^v$ and reference latents $F^r$ are merged into a unified token list, with their information interaction facilitated by self-attention modules in DiT blocks. In this framework, rotary positional encoding (RoPE) \cite{Su2024} plays a crucial role, as it distinguishes video tokens from reference tokens and differentiates between distinct subjects. When extending from single-view to multi-view S2V (MV-S2V), RoPE further needs to discriminate between different subjects and different reference views of the same subject. We conduct a meticulous investigation below into RoPE designs tailored for MV-S2V.

\textbf{Vanilla RoPE.} Following prior works \cite{Jiang2025,Liu2025}, the reference latents are directly appended to video latents along the temporal dimension, as shown in Figure \ref{fig:method} (b). This strategy retains the inherent structure of the base model. However, both different subjects and distinct reference views of the same subject may appear in adjacent frames under this setting, potentially causing the model to confuse these two cases.
 
\textbf{Spatially Shifted RoPE (SS-RoPE).} To avoid such confusion, we consider separating different subjects via frame dimension and distinct views of the same subject via spatial dimension. As shown in Figure \ref{fig:method} (c), we arrange the references of the same subject within a single temporal frame, with different views shifted in the spatial domain. However, such a spatial shift is absent in the base model training and must be learned from scratch during fine-tuning. Additionally, video frames and references lack clear separation in RoPE.

\textbf{Temporally Shifted RoPE (TS-RoPE).} We turn back to a unified discrimination logic within the frame dimension and propose TS-RoPE. As shown in Figure \ref{fig:method} (d), a fixed temporal shift $\delta$ is inserted between the video and references, as well as between the reference latents of different subjects. Different reference views of the same subject are arranged in adjacent frames. This design effectively distinguishes between video frames and references, and achieves clear separation between different subjects and distinct views of the same subject within the references, meanwhile being close to the inherent structure of the base model. The experimental results in Section \ref{sec:expm_abla} consolidate the superiority of this design.

\begin{figure}[t]
\centering
  \includegraphics[width=1.0\linewidth]{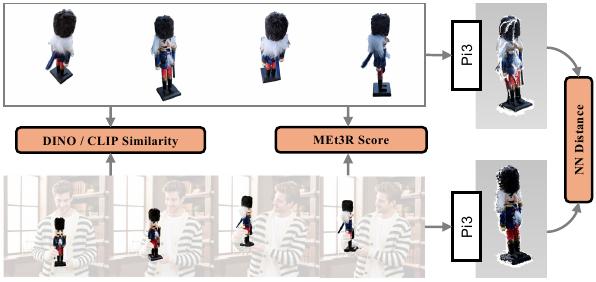}
  \vspace{-0.6cm}
  \caption{Illustration about our multi-view / 3D subject consistency metrics.}
  \vspace{-0.3cm}
  \label{fig:metric}
\end{figure}

\subsection{Training and Inference} 

\textbf{Training Setup.} Our training framework is built upon Rectified Flow (RF) \cite{Lipman2022,Liu2022} with adjusted noise distribution sampling \cite{Esser2024}. The core goal of RF is to learn a flow field capable of transforming Gaussian noise into high-quality, meaningful data samples. During training, the clean video latents $x_0 = F^v$ are first interpolated with Gaussian noise $\epsilon \sim \mathcal{N}(0, I)$ to get the noisy state $x_t = (1 - t) \cdot x_0 + t \cdot \epsilon$, where time step $t$ is randomly sampled and scaled to the range $[0, 1]$ relative to the total diffusion steps ($T=1000$). Our DiT model $G_\theta$ is tasked with predicting the velocity vector $v_t$ to match the true velocity of the interpolation, $u_t = \mathrm{d}x_t/\mathrm{d}t$. The model's prediction is formulated as:
\begin{equation}
    v_t = G_\theta(x_t, t, F^r, y)
\end{equation}
where $F^r$ and $y$ denote reference and textual conditioning. Consequently, the RF training objective is to minimize the Mean Squared Error (MSE) loss between the predicted and ground-truth velocities:
\begin{equation}
    \mathcal{L}_{\text{rf}} = \|v_t - u_t\|^2
\end{equation}

We fine-tune our model from the open-sourced Phantom-Wan model \cite{Phantom-Wan}, which shares the same model architecture with Wan2.1-T2V and has been trained for single-view S2V on large-scale data \cite{Chen2025c}. During training, we apply a 0.1 dropout rate to reference and textual inputs respectively. Furthermore, we randomly drop and shuffle the multi-view reference inputs for each sample to enhance the generalization to varying input view numbers and orders. The model is trained for 2,000 iterations with FusedAdam optimizer (batch size is 64, learning rate is $1 \times 10^{-5}$). The total computational cost amounts to $\sim$ 3,600 GPU hours on A100.

\textbf{Inference Setup.} Denoising is performed with UniPC sampler \cite{Zhao2023} for 50 steps. Classifier-free guidance (CFG) is employed to strengthen both reference and textual conditioning at each step, \ie,
\begin{equation}
    x_{t-1} = x^{\varnothing}_{t-1} + {\omega}_{\boldsymbol{R}} (x^{\boldsymbol{R}}_{t-1} - x^{\varnothing}_{t-1}) + {\omega}_y (x^{\boldsymbol{R},y}_{t-1} - x^{\boldsymbol{R}}_{t-1})
\end{equation}
where $x^{\varnothing}_{t-1}$ denotes the unconditional denoising output, $x^{\boldsymbol{R}}_{t-1}$ denotes the denoising output conditioned on reference images, and $x^{\boldsymbol{R},y}_{t-1}$ denotes the denoising output conditioned on reference images and textual inputs. We set ${\omega}_{\boldsymbol{R}} = 2.5$ and ${\omega}_y = 7.5$.

%% file: sec/5_expm.tex
\section{Experiments}
\label{sec:expm}

\begin{table*}[t]
\centering
\caption{Quantitative results of all methods on Object-Centric (OC) and Human-Object Interaction (HOI) scenes.}
\vspace{-0.3cm}
\resizebox{1.0\linewidth}{!}{
\begin{tabular}{r|cccccc|cc|ccc|c}

\toprule 
\multicolumn{1}{c|}{\multirow{2.5}{*}{\makecell[c]{Object-Centric\\(OC)}}} & \multicolumn{6}{c|}{Multi-View Subject Consistency} & \multicolumn{2}{c|}{3D Subject Consistency} & \multicolumn{3}{c|}{Visual Quality} & \multicolumn{1}{c}{Prompt} \\[+0.1em] 
\cmidrule{2-13} 
& ${S}_{dino}^{v \rightarrow r} \uparrow$ & ${S}_{dino}^{r \rightarrow v} \uparrow$ & ${S}_{clip}^{v \rightarrow r} \uparrow$ & ${S}_{clip}^{r \rightarrow v} \uparrow$ & ${S}_{met3r}^{v \rightarrow r} \downarrow$ & ${S}_{met3r}^{r \rightarrow v} \downarrow$ & ${D}_{nn}^{v \rightarrow r} \downarrow$ & ${D}_{nn}^{r \rightarrow v} \downarrow$ & Aesthetic $\uparrow$ & Imaging $\uparrow$ & Motion $\uparrow$ & ViCLIP $\uparrow$ \\
\midrule
Phantom-SV \cite{Liu2025} & 0.738 & 0.668 & 0.888 & 0.868 & 0.167 & 0.207 & 0.449 & 0.431 & \textbf{0.616} & \textbf{0.747} & \underline{0.994} & \underline{0.237} \\
Phantom-MV \cite{Liu2025} & \underline{0.770} & \underline{0.699} & \textbf{0.907} & \underline{0.887} & \underline{0.151} & 0.192 & \underline{0.168} & \underline{0.212} & 0.596 & 0.704 & \textbf{0.995} & 0.236 \\
MAGREF-SV \cite{Deng2025} & 0.700 & 0.685 & 0.871 & 0.871 & 0.173 & \underline{0.178} & 0.562 & 0.451 & 0.601 & \underline{0.722} & 0.991 & \textbf{0.239} \\
MAGREF-MV \cite{Deng2025} & 0.703 & 0.672 & 0.870 & 0.864 & 0.186 & 0.197 & 0.205 & 0.236 & \underline{0.603} & 0.715 & 0.992 & 0.236 \\
\midrule
\textbf{MV-S2V (Ours)} & \textbf{0.776} & \textbf{0.755} & \underline{0.894} & \textbf{0.893} & \textbf{0.131} & \textbf{0.141} & \textbf{0.110} & \textbf{0.177} & 0.571 & \textbf{0.747} & 0.990 & 0.229 \\
\midrule

\multicolumn{1}{c|}{\multirow{2.5}{*}{\makecell[c]{Human-Object Interaction\\(HOI)}}} & \multicolumn{6}{c|}{Multi-View Subject Consistency} & \multicolumn{2}{c|}{3D Subject Consistency} & \multicolumn{3}{c|}{Visual Quality} & \multicolumn{1}{c}{Prompt} \\[+0.1em] 
\cmidrule{2-13} 
& ${S}_{dino}^{v \rightarrow r} \uparrow$ & ${S}_{dino}^{r \rightarrow v} \uparrow$ & ${S}_{clip}^{v \rightarrow r} \uparrow$ & ${S}_{clip}^{r \rightarrow v} \uparrow$ & ${S}_{met3r}^{v \rightarrow r} \downarrow$ & ${S}_{met3r}^{r \rightarrow v} \downarrow$ & ${D}_{nn}^{v \rightarrow r} \downarrow$ & ${D}_{nn}^{r \rightarrow v} \downarrow$ & Aesthetic $\uparrow$ & Imaging $\uparrow$ & Motion $\uparrow$ & ViCLIP $\uparrow$ \\
\midrule
Phantom-SV \cite{Liu2025} & \underline{0.683} & 0.673 & \underline{0.857} & \underline{0.862} & \textbf{0.171} & 0.183 & 0.337 & 0.312 & \underline{0.587} & \underline{0.752} & \underline{0.993} & 0.187 \\
Phantom-MV \cite{Liu2025} & 0.632 & 0.643 & 0.823 & 0.837 & 0.200 & 0.199 & 0.530 & 0.541 & 0.574 & 0.732 & \underline{0.993} & \underline{0.195} \\
MAGREF-SV \cite{Deng2025} & 0.660 & \textbf{0.701} & 0.832 & 0.859 & 0.191 & 0.188 & \underline{0.325} & \underline{0.268} & 0.549 & 0.741 & 0.983 & 0.177 \\
MAGREF-MV \cite{Deng2025} & 0.646 & 0.679 & 0.823 & 0.848 & 0.198 & \underline{0.181} & 0.588 & 0.570 & 0.561 & 0.732 & 0.984 & 0.189 \\
\midrule
\textbf{MV-S2V (Ours)} & \textbf{0.694} & \underline{0.693} & \textbf{0.858} & \textbf{0.864} & \underline{0.172} & \textbf{0.180} & \textbf{0.247} & \textbf{0.170} & \textbf{0.605} & \textbf{0.761} & \textbf{0.995} & \textbf{0.200} \\
\bottomrule

\end{tabular}
}
\label{tab:exp_main}
\end{table*}

\begin{table*}[h]
\centering
\caption{Ablation study about reference conditioning.}
\vspace{-0.3cm}
\resizebox{1.0\linewidth}{!}{
\begin{tabular}{r|cccccc|cc|cccccc|cc}

\toprule
& \multicolumn{8}{c|}{Object-Centric (OC)} & \multicolumn{8}{c}{Human-Object Interaction (HOI)} \\
\cmidrule{2-17}
& \multicolumn{6}{c|}{Multi-View Subject Consistency} & \multicolumn{2}{c|}{3D Subject Consistency} & \multicolumn{6}{c|}{Multi-View Subject Consistency} & \multicolumn{2}{c}{3D Subject Consistency} \\
\cmidrule{2-17}
& ${S}_{dino}^{v \rightarrow r} \uparrow$ & ${S}_{dino}^{r \rightarrow v} \uparrow$ & ${S}_{clip}^{v \rightarrow r} \uparrow$ & ${S}_{clip}^{r \rightarrow v} \uparrow$ & ${S}_{met3r}^{v \rightarrow r} \downarrow$ & ${S}_{met3r}^{r \rightarrow v} \downarrow$ & ${D}_{nn}^{v \rightarrow r} \downarrow$ & ${D}_{nn}^{r \rightarrow v} \downarrow$ & ${S}_{dino}^{v \rightarrow r} \uparrow$ & ${S}_{dino}^{r \rightarrow v} \uparrow$ & ${S}_{clip}^{v \rightarrow r} \uparrow$ & ${S}_{clip}^{r \rightarrow v} \uparrow$ & ${S}_{met3r}^{v \rightarrow r} \downarrow$ & ${S}_{met3r}^{r \rightarrow v} \downarrow$ & ${D}_{nn}^{v \rightarrow r} \downarrow$ & ${D}_{nn}^{r \rightarrow v} \downarrow$ \\
\midrule
Vanilla & 0.758 & 0.742 & 0.887 & 0.889 & 0.131 & \textbf{0.137} & 0.148 & 0.192 & 0.685 & 0.688 & 0.856 & 0.862 & 0.178 & 0.186 & 0.292 & 0.185 \\
SS-RoPE & 0.765 & 0.742 & 0.891 & 0.889 & 0.131 & 0.145 & 0.662 & 0.601 & 0.672 & 0.675 & 0.853 & 0.857 & 0.182 & 0.190 & 0.278 & 0.178 \\
\midrule
TS-RoPE($\delta$=5) & 0.752 & 0.748 & 0.889 & 0.887 & 0.135 & 0.148 & 0.125 & 0.185 & 0.679 & 0.681 & 0.851 & \textbf{0.866} & 0.174 & 0.182 & 0.263 & \textbf{0.170} \\
\textbf{TS-RoPE($\delta$=10)} & \textbf{0.776} & 0.755 & \textbf{0.894} & \textbf{0.893} & 0.131 & 0.141 & \textbf{0.110} & \textbf{0.177} & \textbf{0.694} & 0.693 & 0.858 & 0.864 & \textbf{0.172} & \textbf{0.180} & \textbf{0.247} & \textbf{0.170} \\
TS-RoPE($\delta$=20) & 0.770 & \textbf{0.759} & 0.891 & 0.891 & \textbf{0.129} & 0.139 & 0.112 & \textbf{0.177} & 0.692 & \textbf{0.696} & \textbf{0.860} & 0.860 & 0.174 & 0.182 & 0.251 & 0.173 \\
\bottomrule

\end{tabular}
}
\vspace{-0.2cm}
\label{tab:abla_rope}
\end{table*}

\subsection{Benchmark} 

We take the 35 objects from NAVI \cite{Jampani2023} dataset and sample 4 sparse views as reference images . We also generate 35 human reference images for HOI scenario with AI image generation tool (Nano-Banana \cite{Banana}) to avoid privacy issues. The 35 sets of multi-view object reference images are then used either independently as input for the OC scenario or combined with human images as input for the HOI scenario, yielding 35 evaluation samples for each scenario. 

\subsection{Evaluation Metrics}

We follow previous works to extensively evaluate S2V generation from three aspects. \textbf{(1) Overall video quality:} we adopt three commomly used metrics from VBench: imaging quality, aesthetic quality, and motion smoothness. \textbf{(2) Text-video consistency:} The prompt following ability is assessed using the ViCLIP score. \textbf{(3) Subject-video consistency:} We detect and segment subjects from videos using Grounded SAM \cite{Ren2024}. After that, we design the following two sets of metrics to measure the consistency between subjects extracted from the video and subjects from reference images, as shown in Figure \ref{fig:metric}.

\textbf{Multi-View Subject Consistency.} For a specific subject, we term the $M$ reference views as $\boldsymbol{I}^r = \{ I^r_m | m=1...M \}$, and the $N$ views from generated video as $\boldsymbol{I}^v = \{ I^v_n | n=1...N \}$. We first measure if each generated view is consistent with at least one of reference views via DINO (or CLIP) feature similarity, \ie,
\begin{equation}
    {S}_{dino}^{v \rightarrow r} = \frac{1}{N} \sum_{n=1}^{N} \max_{m \in \{ 1,...,M \}} {S}_{dino} (I^v_n, I^r_m)
\end{equation}
where ${S}_{dino}(\cdot, \cdot)$ denotes DINO feature similarity between two images. The CLIP-based variant is omitted for brevity.

We also care if generated views can fully cover the provided reference views. Specifically, we measure if each reference view is well displayed by at least one of generated views, \ie, 
\begin{equation}
    {S}_{dino}^{r \rightarrow v} = \frac{1}{M} \sum_{m=1}^{M} \max_{n \in \{ 1,...,N \}} {S}_{dino} (I^v_n, I^r_m)
\end{equation}

Beyond directly measuring the feature similarity based on original images, we also adopt the recently proposed MEt3R score \cite{Asim2025} to measure the feature similarity between the generated views and the reference views with camera viewpoints aligned. Similar to above, we compute MEt3R scores in a bi-directional manner:
\begin{equation}
    {S}_{met3r}^{v \rightarrow r} = \frac{1}{N} \sum_{n=1}^{N} \min_{m \in \{ 1,...,M \}} MEt3R (I^v_n, I^r_m)
\end{equation}
\begin{equation}
    {S}_{met3r}^{r \rightarrow v} = \frac{1}{M} \sum_{m=1}^{M} \min_{n \in \{ 1,...,N \}} MEt3R (I^v_n, I^r_m)
\end{equation}
where $MEt3R(\cdot, \cdot)$ denotes MEt3R score between two images. Readers can refer to \cite{Asim2025} for more details.
 
\textbf{3D Subject Consistency.} We leverage the advanced 3D foundation model $\pi^3$ \cite{Wang2025c} to estimate 3D point clouds $\boldsymbol{P}^r$ and $\boldsymbol{P}^v$ of the subject from reference views and generated views respectively. We first measure if the generated subject point cloud can match at least part of the reference point cloud via nearest-neighbor (NN) distance, \ie,
\begin{equation}
    D_{nn}^{v \rightarrow r} = \frac{1}{|\boldsymbol{P}^v|} \sum_{p^v \in \boldsymbol{P}^v} \min_{p^r \in \boldsymbol{P}^r} {|| p^v - p^r ||}_2
\end{equation}

We also measure if the generated point cloud can fully cover the reference point cloud, \ie,
\begin{equation}
    D_{nn}^{r \rightarrow v} = \frac{1}{|\boldsymbol{P}^r|} \sum_{p^r \in \boldsymbol{P}^r} \min_{p^v \in \boldsymbol{P}^v} {|| p^r - p^v ||}_2
\end{equation}

\subsection{Comparison with Prior Works}

We compare our method with latest open-source methods Phantom \cite{Liu2025} and MAGREF \cite{Deng2025}. Two variants of each baseline are evaluated for fair comparison: \textbf{(1) Single-View (SV):} We follow the common practice of the baselines and feed a single-view reference image for each subject. \textbf{(1) Multi-View (MV):} We feed the baselines with the same multi-view reference images as ours, although such discrepancy between training and inference settings may cause degraded visual quality.

\paragraph{Analysis:} As shown in Table \ref{tab:exp_main}, our method achieves superior performance over the baselines on subject consistency, and competitive performance on visual quality and prompt following ability. As shown in Figures \ref{fig:qual_res1} and \ref{fig:qual_res2}, single-view baselines only adheres to a single reference view and make guesses about other views which are often inconsistent with the actual appearance of real subjects. On the other hand, multi-view baselines tend to generate artifacts such as object deformation or fragmentation, due to discrepancies between the training and inference settings.

\begin{figure*}[t]
\centering
  \includegraphics[width=1.0\linewidth]{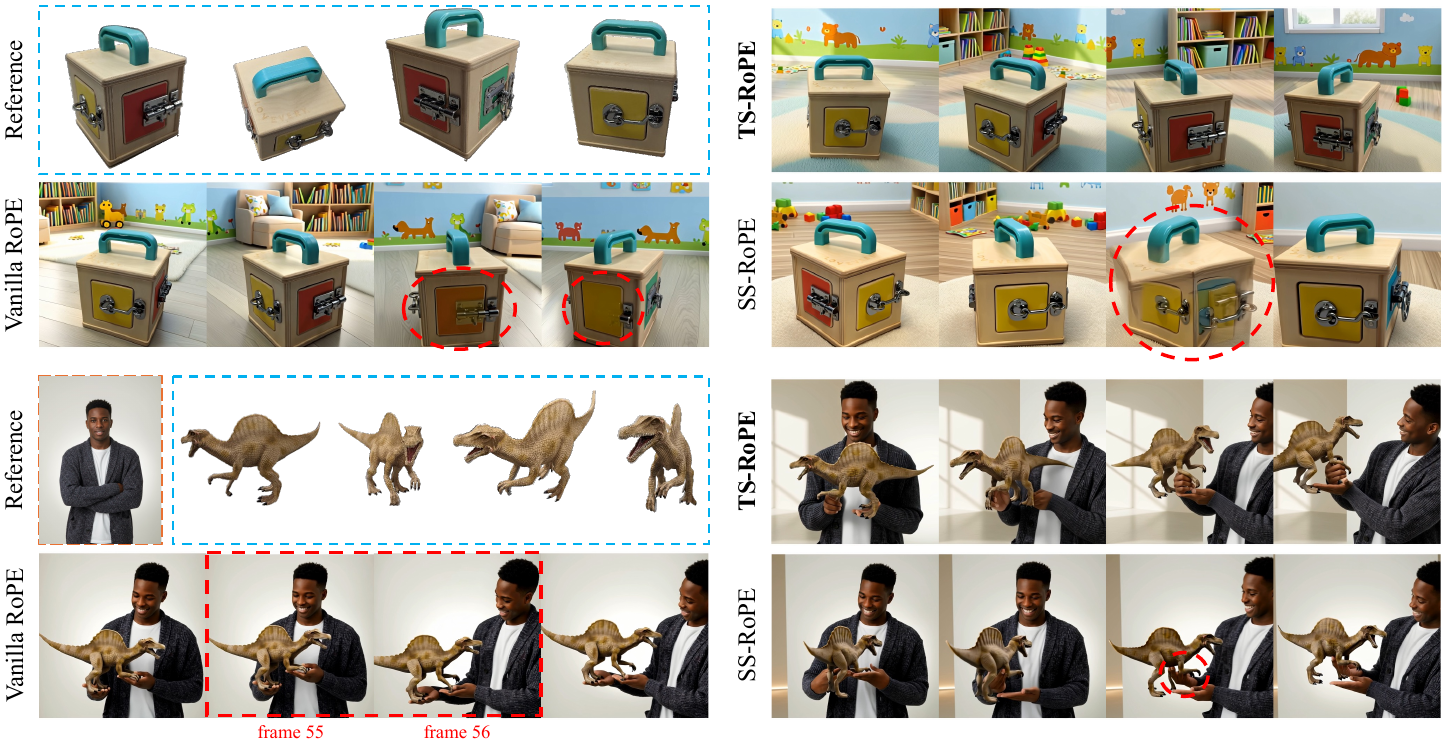}
  \vspace{-0.8cm}
  \caption{Qualitative results of ablation study for reference conditioning. Artifacts in generated results, \ie, object deformation, abrupt changes, are highlighted.}
  \label{fig:abla}
\end{figure*}

\subsection{Ablation Study}
\label{sec:expm_abla}

\textbf{Reference Conditioning.} We ablate reference conditioning designs discussed in Section \ref{sec:rope}: (1) Vanilla RoPE, (2) Spatially Shifted RoPE (SS-RoPE), and (3) Temporally Shifted RoPE (TS-RoPE). As shown in Table \ref{tab:abla_rope}, TS-RoPE achieves the best subject consistency. Figure \ref{fig:abla} further presents qualitative results, where the two sub-optimal designs suffer from object deformation and abrupt content changes, arising from the lack of discrimination between video and cross-view/subject references. 

Furthermore, we analyze the temporal shift $\delta$ in TS-RoPE, testing 5, 10 (default), and 20. Table \ref{tab:abla_rope} shows that a small shift ($\delta=5$) degrades performance, while a larger shift ($\delta=20$) performs similarly to our default ($\delta=10$). This validates that a sufficient temporal shift is crucial for discriminating videos from references.

\textbf{Reference View Numbers.} We further test the applicability to different numbers of reference views. As shown in Table \ref{tab:abla_view}, our method accommodates variable view numbers, while more references tend to yield better multi-view/3D subject consistency.

\textbf{Analysis of Novel View Generalization.} We also quantitatively assess the model's ability to generate novel views using camera-to-subject viewpoint differences. Since quantifying $SE(3)$ pose differences is non-trivial, we propose View Frustum Distance (VFD). We first align the estimated camera poses of references and generated frames into a common coordinate system and normalize the spatial scale. Subsequently, each camera pose is represented by 5 vertices on its unit-scale view frustum, accounting for both position and orientation. VFD is then measured by the distance between corresponding frustum vertices of a generated view and its nearest reference view. A larger VFD indicates a more significantly novel view.

Results in Table \ref{tab:abla_vfd} show our model significantly outperforms the \textit{FrameInterp} baseline (simple interpolation between reference views), confirming that MV-S2V generates substantially diverse new viewpoints. Futhermore, our model consistently demonstrates strong novel view generalization across varying numbers of reference views, even with only a single reference view.

\begin{table*}[t]
\centering
\caption{Ablation study about reference view numbers.}
\vspace{-0.3cm}
\resizebox{1.0\linewidth}{!}{
\begin{tabular}{r|cccccc|cc|cccccc|cc}

\toprule
\multicolumn{1}{c|}{\multirow{4}{*}{\makecell[c]{Number\\of\\Ref. Views}}} & \multicolumn{8}{c|}{Object-Centric (OC)} & \multicolumn{8}{c}{Human-Object Interaction (HOI)} \\
\cmidrule{2-17}
& \multicolumn{6}{c|}{Multi-View Subject Consistency} & \multicolumn{2}{c|}{3D Subject Consistency} & \multicolumn{6}{c|}{Multi-View Subject Consistency} & \multicolumn{2}{c}{3D Subject Consistency} \\
\cmidrule{2-17}
& ${S}_{dino}^{v \rightarrow r} \uparrow$ & ${S}_{dino}^{r \rightarrow v} \uparrow$ & ${S}_{clip}^{v \rightarrow r} \uparrow$ & ${S}_{clip}^{r \rightarrow v} \uparrow$ & ${S}_{met3r}^{v \rightarrow r} \downarrow$ & ${S}_{met3r}^{r \rightarrow v} \downarrow$ & ${D}_{nn}^{v \rightarrow r} \downarrow$ & ${D}_{nn}^{r \rightarrow v} \downarrow$ & ${S}_{dino}^{v \rightarrow r} \uparrow$ & ${S}_{dino}^{r \rightarrow v} \uparrow$ & ${S}_{clip}^{v \rightarrow r} \uparrow$ & ${S}_{clip}^{r \rightarrow v} \uparrow$ & ${S}_{met3r}^{v \rightarrow r} \downarrow$ & ${S}_{met3r}^{r \rightarrow v} \downarrow$ & ${D}_{nn}^{v \rightarrow r} \downarrow$ & ${D}_{nn}^{r \rightarrow v} \downarrow$ \\
\midrule
1 & 0.760 & 0.713 & 0.889 & 0.878 & 0.137 & 0.163 & 0.429 & 0.503 & 0.652 & 0.639 & 0.844 & 0.846 & 0.180 & 0.192 & 0.331 & 0.205 \\
2 & 0.768 & 0.749 & 0.890 & 0.891 & 0.142 & 0.151 & 0.148 & 0.201 & 0.674 & 0.654 & 0.851 & 0.848 & 0.175 & 0.192 & 0.324 & 0.191 \\
3 & 0.775 & 0.752 & \textbf{0.894} & 0.892 & 0.137 & 0.144 & 0.133 & 0.179 & 0.684 & 0.679 & 0.854 & 0.860 & 0.174 & 0.186 & 0.279 & 0.186 \\
4 & \textbf{0.776} & \textbf{0.755} & \textbf{0.894} & \textbf{0.893} & \textbf{0.131} & \textbf{0.141} & \textbf{0.110} & \textbf{0.177} & \textbf{0.694} & \textbf{0.693} & \textbf{0.858} & \textbf{0.864} & \textbf{0.172} & \textbf{0.180} & \textbf{0.247} & \textbf{0.170} \\
\bottomrule

\end{tabular}
}
\vspace{-0.3cm}
\label{tab:abla_view}
\end{table*}

\begin{table}[t]
\centering
\caption{View Frustum Distance (VFD) comparison. Higher is better.}
\vspace{-0.3cm}

\resizebox{\columnwidth}{!}{
\begin{tabular}{r|cc|cc|cccc|c}
\toprule
& \multicolumn{2}{c|}{Phantom} & \multicolumn{2}{c|}{MAGREF} & \multicolumn{4}{c|}{\textbf{MV-S2V (Ours)}} & FrameInterp \\
\midrule
Ref. Views & 1 & 4 & 1 & 4 & 1 & 2 & 3 & 4 & 4 \\
\midrule
OC & 1.211 & 1.096 & \textbf{1.967} & 1.906 & 1.690 & 1.558 & 1.489 & 1.407 & 0.337 \\
HOI & 1.629 & 1.642 & 1.671 & 1.631 & \textbf{2.919} & 2.438 & 2.241 & 1.914 & 0.552 \\
\bottomrule
\end{tabular}
}

\vspace{-0.3cm}
\label{tab:abla_vfd}
\end{table}

%% file: sec/6_sum.tex
\section{Conclusion}
\label{sec:sum}

In this work, we address the limitations of single-view S2V by proposing and solving the Multi-View Subject-to-Video Generation (MV-S2V) task, which enforces 3D subject consistency. To achieve this, we develop a novel framework tackling key issues of data scarcity and multi-view reference conditioning. First, we overcome data scarcity via a highly controllable synthetic data curation pipeline to generate large-scale customized training data, complemented by a small-scale real-world captured dataset. Second, we design an effective TS-RoPE for multi-view reference conditioning, which clearly separates cross-subject and cross-view references in conditional generation. Our framework demonstrates superior performance and remarkable 3D subject consistency, establishing MV-S2V as a crucial next direction for subject-driven video synthesis, especially in high-fidelity applications like advertising and augmented reality.

\textbf{Limitations and future work.} 
In this work, we mainly deal with one central subject with multi-view references in OC scenarios, or with an additional human subject in HOI scenarios. Future works may extend to cases where multiple subjects all have multi-view references. On the other hand, this work focuses on controlling the appearance of a rigid subject in video generation with multi-view references. Future works may extend to controlling a deformable subject with multi-state references, \eg, generating a video of a refrigerator being opened, given reference images of the refrigerator in both closed and open states.

%% file: sec/X_suppl.tex
\section{More Details about Synthetic Data Curation}

\begin{figure*}[b!]
\centering
  \includegraphics[width=1.0\linewidth]{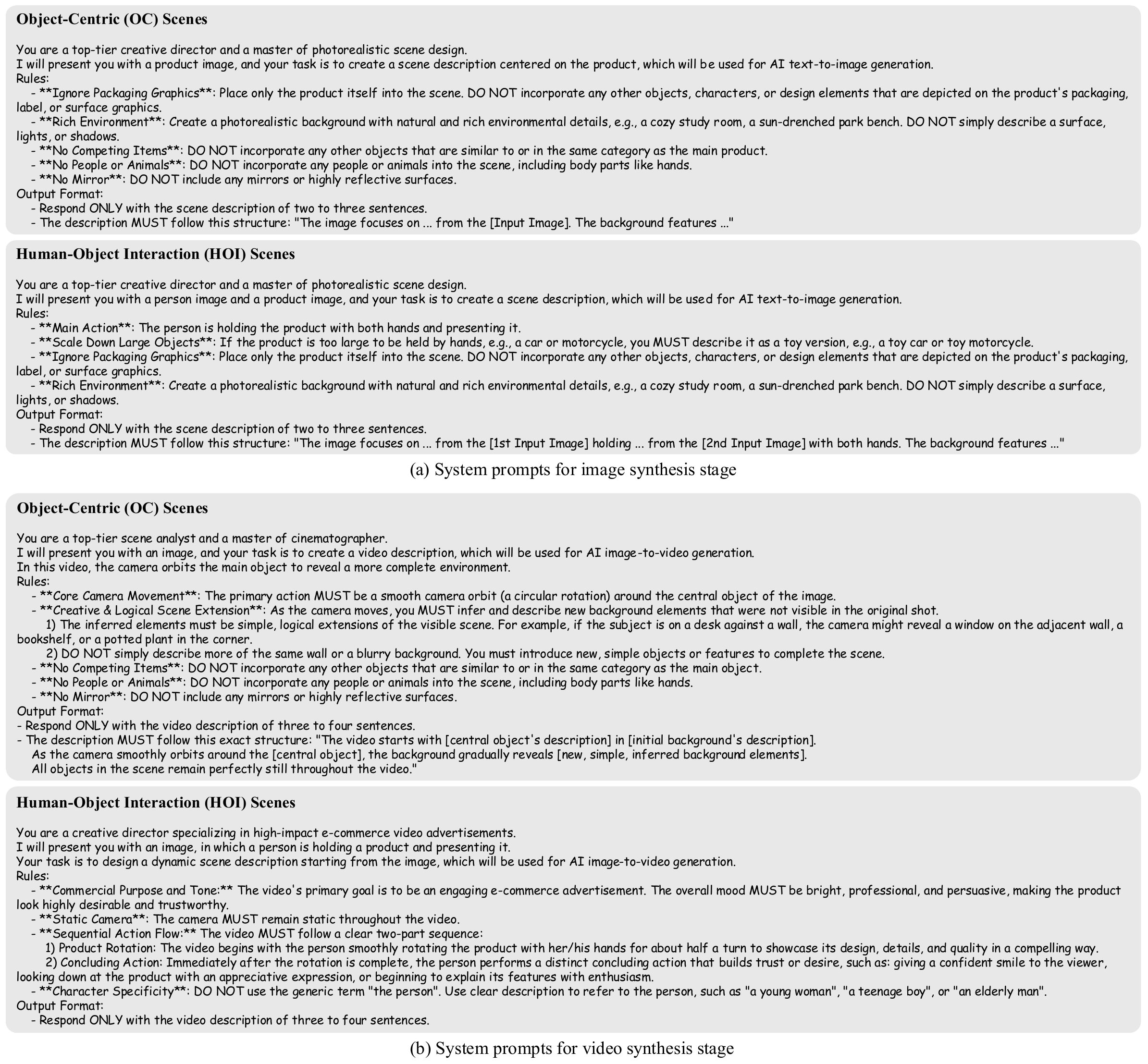}
  \vspace{-0.6cm}
  \caption{System prompts for image and video synthesis stages in our synthetic data curation pipeline.}
\label{fig:prompt1}
\end{figure*}

\begin{figure*}[b!]
\centering
  \includegraphics[width=1.0\linewidth]{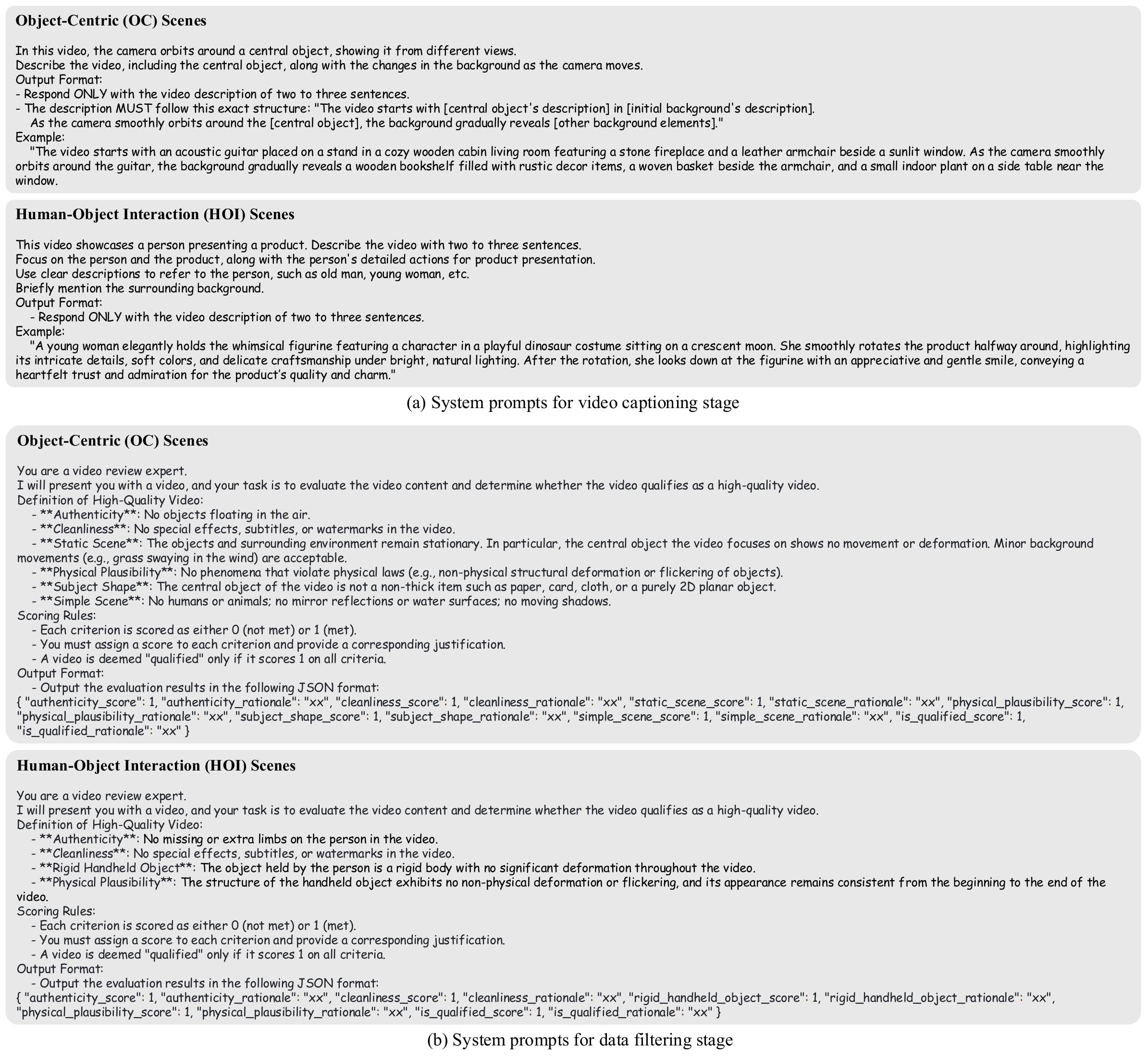}
  \vspace{-0.6cm}
  \caption{System prompts for video captioning and data filtering stages in our synthetic data curation pipeline.}
\label{fig:prompt2}
\end{figure*}

Here we provide additional details about our synthetic data curation. Specifically, we design system prompts for Gemini \cite{Gemini} to generate per-scene text prompts for S2I \cite{Banana} and I2V \cite{Cao2025,Wang2025b} models in image and video synthesis stages. We also design system prompts to guide existing existing VLM \cite{Yuan2025b,Gemini} models to automatically analyze generated video content in the following video captioning and data filtering stages. Figures \ref{fig:prompt1} and \ref{fig:prompt2} illustrates the system prompts employed in each stage of our synthetic data curation pipeline.

\section{More Details about Evaluation Benchmark}

Given object reference images from NAVI \cite{Jampani2023} and generated human reference images, we also employ Gemini \cite{Gemini} to generate per-scene text prompts. The system prompts used are similar to those in video synthesis stage of our synthetic data curation pipeline, so we omit them here for brevity.

%% file: main.bib
@String(CVPR= {IEEE Conf. Comput. Vis. Pattern Recog.})

@String(ICCV= {Int. Conf. Comput. Vis.})

@String(ECCV= {Eur. Conf. Comput. Vis.})

@String(ICLR = {Int. Conf. Learn. Represent.})

@String(AAAI = {AAAI})

@String(CVPR  = {CVPR})

@String(ICCV  = {ICCV})

@String(ECCV  = {ECCV})

@String(ICLR  = {ICLR})

@article{Ho2020,
  author       = {Jonathan Ho and
                  Ajay Jain and
                  Pieter Abbeel},
  title        = {Denoising Diffusion Probabilistic Models},
  journal    = {{NeurIPS}},
  year         = {2020}
}

@article{Rombach2O22,
  author       = {Robin Rombach and
                  Andreas Blattmann and
                  Dominik Lorenz and
                  Patrick Esser and
                  Bj{\"{o}}rn Ommer},
  title        = {High-Resolution Image Synthesis with Latent Diffusion Models},
  journal    = {{CVPR}},
  year         = {2022}
}

@article{Ronneberger2015,
  author       = {Olaf Ronneberger and
                  Philipp Fischer and
                  Thomas Brox},
  title        = {U-Net: Convolutional Networks for Biomedical Image Segmentation},
  journal    = {{MICCAI}},
  year         = {2015}
}

@article{Singer2023,
  author       = {Uriel Singer and
                  Adam Polyak and
                  Thomas Hayes and
                  Xi Yin and
                  Jie An and
                  Songyang Zhang and
                  Qiyuan Hu and
                  Harry Yang and
                  Oron Ashual and
                  Oran Gafni and
                  Devi Parikh and
                  Sonal Gupta and
                  Yaniv Taigman},
  title        = {Make-A-Video: Text-to-Video Generation without Text-Video Data},
  journal    = {{ICLR}},
  year         = {2023}
}

@article{Blattmann2023,
  author       = {Andreas Blattmann and
                  Tim Dockhorn and
                  Sumith Kulal and
                  Daniel Mendelevitch and
                  Maciej Kilian and
                  Dominik Lorenz and
                  Yam Levi and
                  Zion English and
                  Vikram Voleti and
                  Adam Letts and
                  Varun Jampani and
                  Robin Rombach},
  title        = {Stable Video Diffusion: Scaling Latent Video Diffusion Models to Large Datasets},
  journal      = {{arXiv:2311.15127}},
  year         = {2023}
}

@article{Guo2024,
  author       = {Yuwei Guo and
                  Ceyuan Yang and
                  Anyi Rao and
                  Zhengyang Liang and
                  Yaohui Wang and
                  Yu Qiao and
                  Maneesh Agrawala and
                  Dahua Lin and
                  Bo Dai},
  title        = {AnimateDiff: Animate Your Personalized Text-to-Image Diffusion Models without Specific Tuning},
  journal    = {{ICLR}},
  year         = {2024}
}

@article{Peebles2023,
  author       = {William Peebles and
                  Saining Xie},
  title        = {Scalable Diffusion Models with Transformers},
  journal    = {{ICCV}},
  year         = {2023}
}

@article{Esser2024,
  author       = {Patrick Esser and
                  Sumith Kulal and
                  Andreas Blattmann and
                  Rahim Entezari and
                  Jonas M{\"{u}}ller and
                  Harry Saini and
                  Yam Levi and
                  Dominik Lorenz and
                  Axel Sauer and
                  Frederic Boesel and
                  Dustin Podell and
                  Tim Dockhorn and
                  Zion English and
                  Robin Rombach},
  title        = {Scaling Rectified Flow Transformers for High-Resolution Image Synthesis},
  journal    = {{ICML}},
  year         = {2024}
}

@article{Yang2025,
  author       = {Zhuoyi Yang and
                  Jiayan Teng and
                  Wendi Zheng and
                  Ming Ding and
                  Shiyu Huang and
                  Jiazheng Xu and
                  Yuanming Yang and
                  Wenyi Hong and
                  Xiaohan Zhang and
                  Guanyu Feng and
                  Da Yin and
                  Yuxuan Zhang and
                  Weihan Wang and
                  Yean Cheng and
                  Bin Xu and
                  Xiaotao Gu and
                  Yuxiao Dong and
                  Jie Tang},
  title        = {CogVideoX: Text-to-Video Diffusion Models with An Expert Transformer},
  journal    = {{ICLR}},
  year         = {2025}
}

@article{Kong2024,
  author       = {Weijie Kong and
                  Qi Tian and
                  Zijian Zhang and
                  Rox Min and
                  Zuozhuo Dai and
                  Jin Zhou and
                  Jiangfeng Xiong and
                  Xin Li and
                  Bo Wu and
                  Jianwei Zhang and
                  Kathrina Wu and
                  Qin Lin and
                  Junkun Yuan and
                  Yanxin Long and
                  Aladdin Wang and
                  Andong Wang and
                  Changlin Li and
                  Duojun Huang and
                  Fang Yang and
                  Hao Tan and
                  Hongmei Wang and
                  Jacob Song and
                  Jiawang Bai and
                  Jianbing Wu and
                  Jinbao Xue and
                  Joey Wang and
                  Kai Wang and
                  Mengyang Liu and
                  Pengyu Li and
                  Shuai Li and
                  Weiyan Wang and
                  Wenqing Yu and
                  Xinchi Deng and
                  Yang Li and
                  Yi Chen and
                  Yutao Cui and
                  Yuanbo Peng and
                  Zhentao Yu and
                  Zhiyu He and
                  Zhiyong Xu and
                  Zixiang Zhou and
                  Zunnan Xu and
                  Yangyu Tao and
                  Qinglin Lu and
                  Songtao Liu and
                  Daquan Zhou and
                  Hongfa Wang and
                  Yong Yang and
                  Di Wang and
                  Yuhong Liu and
                  Jie Jiang and
                  Caesar Zhong},
  title        = {HunyuanVideo: {A} Systematic Framework For Large Video Generative Models},
  journal      = {{arXiv:2412.03603}},
  year         = {2024}
}

@article{Wang2025,
  author       = {Jianyi Wang and
                  Zhijie Lin and
                  Meng Wei and
                  Yang Zhao and
                  Ceyuan Yang and
                  Chen Change Loy and
                  Lu Jiang},
  title        = {SeedVR: Seeding Infinity in Diffusion Transformer Towards Generic Video Restoration},
  journal    = {{CVPR}},
  year         = {2025}
}

@article{Gal2023,
  author       = {Rinon Gal and
                  Yuval Alaluf and
                  Yuval Atzmon and
                  Or Patashnik and
                  Amit Haim Bermano and
                  Gal Chechik and
                  Daniel Cohen{-}Or},
  title        = {An Image is Worth One Word: Personalizing Text-to-Image Generation using Textual Inversion},
  journal    = {{ICLR}},
  year         = {2023}
}

@article{Hu2022,
  author       = {Edward J. Hu and
                  Yelong Shen and
                  Phillip Wallis and
                  Zeyuan Allen{-}Zhu and
                  Yuanzhi Li and
                  Shean Wang and
                  Lu Wang and
                  Weizhu Chen},
  title        = {LoRA: Low-Rank Adaptation of Large Language Models},
  journal    = {{ICLR}},
  year         = {2022}
}

@article{Huang2024,
  author       = {Lianghua Huang and
                  Wei Wang and
                  Zhi{-}Fan Wu and
                  Yupeng Shi and
                  Huanzhang Dou and
                  Chen Liang and
                  Yutong Feng and
                  Yu Liu and
                  Jingren Zhou},
  title        = {In-Context LoRA for Diffusion Transformers},
  journal      = {{arXiv:2410.23775}},
  year         = {2024}
}

@article{Ruiz2023,
  author       = {Nataniel Ruiz and
                  Yuanzhen Li and
                  Varun Jampani and
                  Yael Pritch and
                  Michael Rubinstein and
                  Kfir Aberman},
  title        = {DreamBooth: Fine Tuning Text-to-Image Diffusion Models for Subject-Driven Generation},
  journal    = {{CVPR}},
  year         = {2023}
}

@article{Shah2024,
  author       = {Viraj Shah and
                  Nataniel Ruiz and
                  Forrester Cole and
                  Erika Lu and
                  Svetlana Lazebnik and
                  Yuanzhen Li and
                  Varun Jampani},
  title        = {ZipLoRA: Any Subject in Any Style by Effectively Merging LoRAs},
  journal    = {{ECCV}},
  year         = {2024}
}

@article{Ye2023,
  author       = {Hu Ye and
                  Jun Zhang and
                  Sibo Liu and
                  Xiao Han and
                  Wei Yang},
  title        = {IP-Adapter: Text Compatible Image Prompt Adapter for Text-to-Image Diffusion Models},
  journal      = {{arXiv:2308.06721}},
  year         = {2023}
}

@article{Chen2024,
  author       = {Zhuowei Chen and
                  Shancheng Fang and
                  Wei Liu and
                  Qian He and
                  Mengqi Huang and
                  Zhendong Mao},
  title        = {DreamIdentity: Enhanced Editability for Efficient Face-Identity Preserved Image Generation},
  journal    = {{AAAI}},
  year         = {2024}
}

@article{Guo2024b,
  author       = {Zinan Guo and
                  Yanze Wu and
                  Zhuowei Chen and
                  Lang Chen and
                  Peng Zhang and
                  Qian He},
  title        = {PuLID: Pure and Lightning {ID} Customization via Contrastive Alignment},
  journal    = {{NeurIPS}},
  year         = {2024}
}

@article{Wang2024,
  author       = {Qixun Wang and
                  Xu Bai and
                  Haofan Wang and
                  Zekui Qin and
                  Anthony Chen},
  title        = {InstantID: Zero-shot Identity-Preserving Generation in Seconds},
  journal      = {{arXiv:2401.07519}},
  year         = {2024}
}

@article{Cherti2023,
  author       = {Mehdi Cherti and
                  Romain Beaumont and
                  Ross Wightman and
                  Mitchell Wortsman and
                  Gabriel Ilharco and
                  Cade Gordon and
                  Christoph Schuhmann and
                  Ludwig Schmidt and
                  Jenia Jitsev},
  title        = {Reproducible Scaling Laws for Contrastive Language-Image Learning},
  journal    = {{CVPR}},
  year         = {2023}
}

@article{Oquab2024,
  author       = {Maxime Oquab and
                  Timoth{\'{e}}e Darcet and
                  Th{\'{e}}o Moutakanni and
                  Huy V. Vo and
                  Marc Szafraniec and
                  Vasil Khalidov and
                  Pierre Fernandez and
                  Daniel Haziza and
                  Francisco Massa and
                  Alaaeldin El{-}Nouby and
                  Mido Assran and
                  Nicolas Ballas and
                  Wojciech Galuba and
                  Russell Howes and
                  Po{-}Yao Huang and
                  Shang{-}Wen Li and
                  Ishan Misra and
                  Michael Rabbat and
                  Vasu Sharma and
                  Gabriel Synnaeve and
                  Hu Xu and
                  Herv{\'{e}} J{\'{e}}gou and
                  Julien Mairal and
                  Patrick Labatut and
                  Armand Joulin and
                  Piotr Bojanowski},
  title        = {DINOv2: Learning Robust Visual Features without Supervision},
  journal      = {{Trans. Mach. Learn. Res.}},
  year         = {2024}
}

@article{Chen2025,
  author       = {Xi Chen and
                  Zhifei Zhang and
                  He Zhang and
                  Yuqian Zhou and
                  Soo Ye Kim and
                  Qing Liu and
                  Yijun Li and
                  Jianming Zhang and
                  Nanxuan Zhao and
                  Yilin Wang and
                  Hui Ding and
                  Zhe Lin and
                  Hengshuang Zhao},
  title        = {UniReal: Universal Image Generation and Editing via Learning Real-world
                  Dynamics},
  journal    = {{CVPR}},
  year         = {2025}
}

@article{Han2025,
  author       = {Zhen Han and
                  Zeyinzi Jiang and
                  Yulin Pan and
                  Jingfeng Zhang and
                  Chaojie Mao and
                  Chen{-}Wei Xie and
                  Yu Liu and
                  Jingren Zhou},
  title        = {{ACE:} All-round Creator and Editor Following Instructions via Diffusion Transformer},
  journal    = {{ICLR}},
  year         = {2025}
}

@article{Xiao2025,
  author       = {Shitao Xiao and
                  Yueze Wang and
                  Junjie Zhou and
                  Huaying Yuan and
                  Xingrun Xing and
                  Ruiran Yan and
                  Chaofan Li and
                  Shuting Wang and
                  Tiejun Huang and
                  Zheng Liu},
  title        = {OmniGen: Unified Image Generation},
  journal    = {{CVPR}},
  year         = {2025}
}

@misc{Sora,
  author       = {OpenAI},
  title        = {Sora},
  howpublished = {\url{https://openai.com}},
  year         = {2023},
}

@misc{Gemini,
  author       = {Google},
  title        = {Gemini},
  howpublished = {\url{https://gemini.google.com}},
  year         = {2025},
}

@misc{Banana,
  author       = {Google},
  title        = {Nano Banana},
  howpublished = {\url{https://aistudio.google.com/models/gemini-2-5-flash-image}},
  year         = {2025},
}

@misc{Kling,
  author       = {Kling},
  title        = {Image to Video},
  howpublished = {\url{https://app.klingai.com/global/image-to-video/multi-id/new}},
  year         = {2024},
}

@misc{Phantom-Wan,
  author       = {HuggingFace},
  title        = {Phantom-Wan},
  howpublished = {\url{https://huggingface.co/bytedance-research/Phantom}},
  year         = {2025},
}

@article{He2024,
  author       = {Xuanhua He and
                  Quande Liu and
                  Shengju Qian and
                  Xin Wang and
                  Tao Hu and
                  Ke Cao and
                  Keyu Yan and
                  Man Zhou and
                  Jie Zhang},
  title        = {ID-Animator: Zero-Shot Identity-Preserving Human Video Generation},
  journal      = {{arXiv:2404.15275}},
  year         = {2024}
}

@article{Yuan2025,
  author       = {Shenghai Yuan and
                  Jinfa Huang and
                  Xianyi He and
                  Yunyang Ge and
                  Yujun Shi and
                  Liuhan Chen and
                  Jiebo Luo and
                  Li Yuan},
  title        = {Identity-Preserving Text-to-Video Generation by Frequency Decomposition},
  journal    = {{CVPR}},
  year         = {2025}
}

@article{Huang2025,
  author       = {Yuzhou Huang and
                  Ziyang Yuan and
                  Quande Liu and
                  Qiulin Wang and
                  Xintao Wang and
                  Ruimao Zhang and
                  Pengfei Wan and
                  Di Zhang and
                  Kun Gai},
  title        = {ConceptMaster: Multi-Concept Video Customization on Diffusion Transformer Models Without Test-Time Tuning},
  journal      = {{arXiv:2501.04698}},
  year         = {2025},
}

@article{Liang2025,
  author       = {Feng Liang and
                  Haoyu Ma and
                  Zecheng He and
                  Tingbo Hou and
                  Ji Hou and
                  Kunpeng Li and
                  Xiaoliang Dai and
                  Felix Juefei{-}Xu and
                  Samaneh Azadi and
                  Animesh Sinha and
                  Peizhao Zhang and
                  Peter Vajda and
                  Diana Marculescu},
  title        = {Movie Weaver: Tuning-Free Multi-Concept Video Personalization with Anchored Prompts},
  journal    = {{CVPR}},
  year         = {2025}
}

@article{Chen2025b,
  author       = {Tsai{-}Shien Chen and
                  Aliaksandr Siarohin and
                  Willi Menapace and
                  Yuwei Fang and
                  Kwot Sin Lee and
                  Ivan Skorokhodov and
                  Kfir Aberman and
                  Jun{-}Yan Zhu and
                  Ming{-}Hsuan Yang and
                  Sergey Tulyakov},
  title        = {Multi-subject Open-set Personalization in Video Generation},
  journal    = {{CVPR}},
  year         = {2025}
}

@article{Jiang2025,
  author       = {Zeyinzi Jiang and
                  Zhen Han and
                  Chaojie Mao and
                  Jingfeng Zhang and
                  Yulin Pan and
                  Yu Liu},
  title        = {{VACE:} All-in-One Video Creation and Editing},
  journal    = {{ICCV}},
  year         = {2025}
}

@article{Liu2025,
  author       = {Lijie Liu and
                  Tianxiang Ma and
                  Bingchuan Li and
                  Zhuowei Chen and
                  Jiawei Liu and
                  Qian He and
                  Xinglong Wu},
  title        = {Phantom: Subject-consistent video generation via cross-modal alignment},
  journal    = {{ICCV}},
  year         = {2025}
}

@article{Deng2025,
  author       = {Yufan Deng and
                  Xun Guo and
                  Yuanyang Yin and
                  Jacob Zhiyuan Fang and
                  Yiding Yang and
                  Yizhi Wang and
                  Shenghai Yuan and
                  Angtian Wang and
                  Bo Liu and
                  Haibin Huang and
                  Chongyang Ma},
  title        = {{MAGREF:} Masked Guidance for Any-Reference Video Generation},
  journal      = {{arXiv:2505.23742}},
  year         = {2025}
}

@article{Zhang2025,
  author       = {Zhenxing Zhang and
                  Jiayan Teng and
                  Zhuoyi Yang and
                  Tiankun Cao and
                  Cheng Wang and
                  Xiaotao Gu and
                  Jie Tang and
                  Dan Guo and
                  Meng Wang},
  title        = {Kaleido: Open-Sourced Multi-Subject Reference Video Generation Model},
  journal      = {{arXiv:2510.18573}},
  year         = {2025}
}

@article{Cao2025,
  author       = {Chenjie Cao and
                  Jingkai Zhou and
                  Shikai Li and
                  Jingyun Liang and
                  Chaohui Yu and
                  Fan Wang and
                  Xiangyang Xue and
                  Yanwei Fu},
  title        = {Uni3C: Unifying Precisely 3D-Enhanced Camera and Human Motion Controls for Video Generation},
  journal      = {{arXiv:2504.14899}},
  year         = {2025}
}

@article{Wang2025b,
  author       = {Ang Wang and
                  Baole Ai and
                  Bin Wen and
                  Chaojie Mao and
                  Chen{-}Wei Xie and
                  Di Chen and
                  Feiwu Yu and
                  Haiming Zhao and
                  Jianxiao Yang and
                  Jianyuan Zeng and
                  Jiayu Wang and
                  Jingfeng Zhang and
                  Jingren Zhou and
                  Jinkai Wang and
                  Jixuan Chen and
                  Kai Zhu and
                  Kang Zhao and
                  Keyu Yan and
                  Lianghua Huang and
                  Xiaofeng Meng and
                  Ningyi Zhang and
                  Pandeng Li and
                  Pingyu Wu and
                  Ruihang Chu and
                  Ruili Feng and
                  Shiwei Zhang and
                  Siyang Sun and
                  Tao Fang and
                  Tianxing Wang and
                  Tianyi Gui and
                  Tingyu Weng and
                  Tong Shen and
                  Wei Lin and
                  Wei Wang and
                  Wei Wang and
                  Wenmeng Zhou and
                  Wente Wang and
                  Wenting Shen and
                  Wenyuan Yu and
                  Xianzhong Shi and
                  Xiaoming Huang and
                  Xin Xu and
                  Yan Kou and
                  Yangyu Lv and
                  Yifei Li and
                  Yijing Liu and
                  Yiming Wang and
                  Yingya Zhang and
                  Yitong Huang and
                  Yong Li and
                  You Wu and
                  Yu Liu and
                  Yulin Pan and
                  Yun Zheng and
                  Yuntao Hong and
                  Yupeng Shi and
                  Yutong Feng and
                  Zeyinzi Jiang and
                  Zhen Han and
                  Zhi{-}Fan Wu and
                  Ziyu Liu},
  title        = {Wan: Open and Advanced Large-Scale Video Generative Models},
  journal      = {{arXiv:2503.20314}},
  year         = {2025}
}

@article{Yuan2025b,
  author       = {Liping Yuan and
                  Jiawei Wang and
                  Haomiao Sun and
                  Yuchen Zhang and
                  Yuan Lin},
  title        = {Tarsier2: Advancing Large Vision-Language Models from Detailed Video Description to Comprehensive Video Understanding},
  journal      = {{arXiv:2501.07888}},
  year         = {2025}
}

@article{Reizenstein2021,
  author       = {Jeremy Reizenstein and
                  Roman Shapovalov and
                  Philipp Henzler and
                  Luca Sbordone and
                  Patrick Labatut and
                  David Novotn{\'{y}}},
  title        = {Common Objects in 3D: Large-Scale Learning and Evaluation of Real-life 3D Category Reconstruction},
  journal    = {{ICCV}},
  year         = {2021}
}

@article{Liu2025b,
  author       = {Kun Liu and
                  Qi Liu and
                  Xinchen Liu and
                  Jie Li and
                  Yongdong Zhang and
                  Jiebo Luo and
                  Xiaodong He and
                  Wu Liu},
  title        = {HOIGen-1M: {A} Large-scale Dataset for Human-Object Interaction Video Generation},
  journal    = {{CVPR}},
  year         = {2025}
}

@article{Chen2025c,
  author       = {Zhuowei Chen and
                  Bingchuan Li and
                  Tianxiang Ma and
                  Lijie Liu and
                  Mingcong Liu and
                  Yi Zhang and
                  Gen Li and
                  Xinghui Li and
                  Siyu Zhou and
                  Qian He and
                  Xinglong Wu},
  title        = {Phantom-Data : Towards a General Subject-Consistent Video Generation Dataset},
  journal      = {{arXiv:2506.18851}},
  year         = {2025}
}

@article{Jampani2023,
  author       = {Varun Jampani and
                  Kevis{-}Kokitsi Maninis and
                  Andreas Engelhardt and
                  Arjun Karpur and
                  Karen Truong and
                  Kyle Sargent and
                  Stefan Popov and
                  Andr{\'{e}} Ara{\'{u}}jo and
                  Ricardo Martin{-}Brualla and
                  Kaushal Patel and
                  Daniel Vlasic and
                  Vittorio Ferrari and
                  Ameesh Makadia and
                  Ce Liu and
                  Yuanzhen Li and
                  Howard Zhou},
  title        = {{NAVI:} Category-Agnostic Image Collections with High-Quality 3D Shape and Pose Annotations},
  journal    = {{NeurIPS}},
  year         = {2023}
}

@article{Wang2025c,
  author       = {Yifan Wang and
                  Jianjun Zhou and
                  Haoyi Zhu and
                  Wenzheng Chang and
                  Yang Zhou and
                  Zizun Li and
                  Junyi Chen and
                  Jiangmiao Pang and
                  Chunhua Shen and
                  Tong He},
  title        = {{\(\pi\)}\({}^{\mbox{3}}\): Scalable Permutation-Equivariant Visual Geometry Learning},
  journal      = {{arXiv:2507.13347}},
  year         = {2025}
}

@article{Ren2024,
  author       = {Tianhe Ren and
                  Shilong Liu and
                  Ailing Zeng and
                  Jing Lin and
                  Kunchang Li and
                  He Cao and
                  Jiayu Chen and
                  Xinyu Huang and
                  Yukang Chen and
                  Feng Yan and
                  Zhaoyang Zeng and
                  Hao Zhang and
                  Feng Li and
                  Jie Yang and
                  Hongyang Li and
                  Qing Jiang and
                  Lei Zhang},
  title        = {Grounded SAM: Assembling Open-World Models for Diverse Visual Tasks},
  journal      = {{arXiv:2401.14159}},
  year         = {2024}
}

@article{Raffel2019,
  author       = {Colin Raffel and
                  Noam M. Shazeer and
                  Adam Roberts and
                  Katherine Lee and
                  Sharan Narang and
                  Michael Matena and
                  Yanqi Zhou and
                  Wei Li and
                  Peter J. Liu},
  title        = {Exploring the Limits of Transfer Learning with a Unified Text-to-Text Transformer},
  journal      = {J. Mach. Learn. Res.},
  year         = {2019}
}

@article{Lipman2022,
  author       = {Yaron Lipman and
                  Ricky T. Q. Chen and
                  Heli Ben-Hamu and
                  Maximilian Nickel and
                  Matt Le},
  title        = {Flow Matching for Generative Modeling},
  journal    = {{ICLR}},
  year         = {2022}
}

@article{Liu2022,
  author       = {Xingchao Liu and
                  Chengyue Gong and
                  Qiang Liu},
  title        = {Flow Straight and Fast: Learning to Generate and Transfer Data with Rectified Flow},
  journal    = {{ICLR}},
  year         = {2022}
}

@article{Zhao2023,
  author    = {Wenliang Zhao and
               Lujia Bai and
               Yongming Rao and
               Jie Zhou and
               Jiwen Lu},
  title     = {UniPC: A Unified Predictor-Corrector Framework for Fast Sampling of Diffusion Models},
  journal = {{NeurIPS}},
  year      = {2023}
}

@article{Zhao2025,
  title={CETCAM: Camera-Controllable Video Generation via Consistent and Extensible Tokenization},
  author={Zhao, Zelin and Gong, Xinyu and Liu, Bangya and Song, Ziyang and Zhang, Jun and Wu, Suhui and Chen, Yongxin and Zhang, Hao},
  journal={{arXiv:2512.19020}},
  year={2025}
}

@article{Liu2025c,
  title={ByteLoom: Weaving Geometry-Consistent Human-Object Interactions through Progressive Curriculum Learning},
  author={Liu, Bangya and Gong, Xinyu and Zhao, Zelin and Song, Ziyang and Lu, Yulei and Wu, Suhui and Zhang, Jun and Banerjee, Suman and Zhang, Hao},
  journal={{arXiv:2512.22854}},
  year={2025}
}

@article{Asim2025,
  author       = {Mohammad Asim and
                  Christopher Wewer and
                  Thomas Wimmer and
                  Bernt Schiele and
                  Jan Eric Lenssen},
  title        = {{MET3R:} Measuring Multi-View Consistency in Generated Images},
  journal      = {{CVPR}},
  year         = {2025}
}

@article{Su2024,
  author       = {Jianlin Su and
                  Murtadha H. M. Ahmed and
                  Yu Lu and
                  Shengfeng Pan and
                  Wen Bo and
                  Yunfeng Liu},
  title        = {RoFormer: Enhanced transformer with Rotary Position Embedding},
  journal      = {Neurocomputing},
  year         = {2024}
}

@article{Wu2025,
  author       = {Chenfei Wu and
                  Jiahao Li and
                  Jingren Zhou and
                  Junyang Lin and
                  Kaiyuan Gao and
                  Kun Yan and
                  Shengming Yin and
                  Shuai Bai and
                  Xiao Xu and
                  Yilei Chen and
                  Yuxiang Chen and
                  Zecheng Tang and
                  Zekai Zhang and
                  Zhengyi Wang and
                  An Yang and
                  Bowen Yu and
                  Chen Cheng and
                  Dayiheng Liu and
                  Deqing Li and
                  Hang Zhang and
                  Hao Meng and
                  Hu Wei and
                  Jingyuan Ni and
                  Kai Chen and
                  Kuan Cao and
                  Liang Peng and
                  Lin Qu and
                  Minggang Wu and
                  Peng Wang and
                  Shuting Yu and
                  Tingkun Wen and
                  Wensen Feng and
                  Xiaoxiao Xu and
                  Yi Wang and
                  Yichang Zhang and
                  Yongqiang Zhu and
                  Yujia Wu and
                  Yuxuan Cai and
                  Zenan Liu},
  title        = {Qwen-Image Technical Report},
  journal      = {arXiv:2508.02324},
  year         = {2025}
}

@article{Zhang2025a,
  author       = {Jiexuan Zhang and
                  Yiheng Du and
                  Qian Wang and
                  Weiqi Li and
                  Yu Gu and
                  Jian Zhang},
  title        = {AlignedGen: Aligning Style Across Generated Images},
  journal      = {{NeurIPS}},
  year         = {2025}
}

@article{Bai2026,
  author       = {Yunpeng Bai and
                  Haoxiang Li and
                  Qixing Huang},
  title        = {Positional Encoding Field},
  journal      = {{ICLR}},
  year         = {2026}
}

@article{Wu2025a,
  author       = {Ziyi Wu and
                  Aliaksandr Siarohin and
                  Willi Menapace and
                  Ivan Skorokhodov and
                  Yuwei Fang and
                  Varnith Chordia and
                  Igor Gilitschenski and
                  Sergey Tulyakov},
  title        = {Mind the Time: Temporally-Controlled Multi-Event Video Generation},
  journal      = {{CVPR}},
  year         = {2025}
}

@article{Guo2024a,
  author       = {Zinan Guo and
                  Yanze Wu and
                  Zhuowei Chen and
                  Lang Chen and
                  Peng Zhang and
                  Qian He},
  title        = {PuLID: Pure and Lightning ID Customization via Contrastive Alignment},
  journal      = {{NeurIPS}},
  year         = {2024}
}

@article{Qian2025,
  author       = {Guocheng Qian and
                  Kuan-Chieh Wang and
                  Or Patashnik and
                  Negin Heravi and
                  Daniil Ostashev and
                  Sergey Tulyakov and
                  Daniel Cohen-Or and
                  Kfir Aberman},
  title        = {Omni-ID: Holistic Identity Representation Designed for Generative Tasks},
  journal      = {{CVPR}},
  year         = {2025}
}

@article{Liu2023Zero123,
  author       = {Chuanxia Zheng and
                  Long Lian and
                  Yijun Li and
                  Jingyi Yu and
                  Trevor Darrell and
                  Eli Shechtman and
                  Richard Zhang},
  title        = {Zero-1-to-3: Zero-shot One Image to 3D Object},
  journal      = {{ICCV}},
  year         = {2023}
}

@article{Zhou2024SyncDreamer,
  author       = {Yuanxun Lu and
                  Yiheng Du and
                  Chao Xu and
                  Hang Zhou and
                  Tianshu Hu and
                  Lin Ma and
                  Hao Zhao},
  title        = {SyncDreamer: Generating Multiview-consistent Images from a Single-view Image},
  journal      = {{CVPR}},
  year         = {2024}
}

@article{Shi2023MVDream,
  author       = {Yichun Shi and
                  Peng Wang and
                  Jianglong Ye and
                  Yang Xiao and
                  Long Lian and
                  Yijun Li},
  title        = {MVDream: Multi-view Diffusion for 3D Generation},
  journal      = {{NeurIPS}},
  year         = {2023}
}

@article{Voleti2024SV3D,
  author       = {Vikram Voleti and
                  Chun-Han Yao and
                  Mark Boss and
                  Adam W. Harley and
                  Leonid Sigal and
                  Christian Theobalt and
                  Varun Jampani},
  title        = {SV3D: Novel Multi-view Synthesis and 3D Generation from a Single Image using Latent Video Diffusion},
  journal      = {{arXiv:2403.12008}},
  year         = {2024}
}

@article{Tang2024LGM,
  author       = {Ruoxi Tang and
                  Junfeng Yang and
                  Yuxue Yang and
                  Hongyu Yang and
                  Peng Wang and
                  Yichun Shi},
  title        = {LGM: Large Multi-View Gaussian Model for High-Fidelity 3D Generation},
  journal      = {{SIGGRAPH Asia}},
  year         = {2024}
}

@article{Xiang2025SLAT,
  author       = {Jianfeng Xiang and
                  Zelong Lv and
                  Sicheng Xu and
                  Yu Deng and
                  Ruicheng Wang and
                  Bowen Zhang and
                  Dong Chen and
                  Xin Tong and
                  Jiaolong Yang},
  title        = {Structured 3D Latents for Scalable and Versatile 3D Generation},
  journal      = {{CVPR}},
  year         = {2025}
}
